%% file: Main.tex
\documentclass[11pt,letterpaper]{article}
\input{preamble}

\usepackage{emnlp2017}
\usepackage{times}
\usepackage{latexsym}
\usepackage{standalone}
\usepackage{subfiles}
\usepackage{bm}
\usepackage{multirow}

\emnlpfinalcopy

\def\emnlppaperid{1260}

\title{A causal framework for explaining the predictions of \\ black-box sequence-to-sequence models}

\author{David Alvarez-Melis \and Tommi S. Jaakkola \\
	CSAIL, MIT \\ {\tt \{davidam, tommi\}@csail.mit.edu}}
\date{}

\def\biblio{\bibliographystyle{emnlp_natbib}\bibliography{InterpSeq.bib}}

\begin{document}
\def\biblio{}

\maketitle

\begin{abstract}
We interpret the predictions of any black-box structured input-structured output model around a specific input-output pair. Our method returns an ``explanation'' consisting of groups of input-output tokens that are causally related. These dependencies are inferred by querying the black-box model with perturbed inputs, generating a graph over tokens from the responses, and solving a partitioning problem to select the most relevant components. We focus the general approach on sequence-to-sequence problems, adopting a variational autoencoder to yield meaningful input perturbations. We test our method across several NLP sequence generation tasks.
\end{abstract}

\section{Introduction}

Interpretability is often the first casualty when adopting complex predictors. This is particularly true for structured prediction methods at the core of many natural language processing tasks such as machine translation (MT). For example, deep learning models for NLP involve a large number of parameters and complex architectures, making them practically black-box systems. While such systems achieve state-of-the-art results in MT \citep{Bahdanau2014Neural}, summarization \citep{Rush2015Neural} and speech recognition \citep{Chan2015Listen}, they remain largely uninterpretable, although attention mechanisms \cite{Bahdanau2014Neural} can shed some light on how they operate. 

Stronger forms of interpretability could offer several advantages, from trust in model predictions, error analysis, to model refinement. For example, critical medical decisions are increasingly being assisted by complex predictions that should lend themselves to easy verification by human experts. Without understanding how inputs get mapped to the outputs, it is also challenging to diagnose the source of potential errors. A slightly less obvious application concerns model improvement \citep{Ribeiro2016Why} where interpretability can be used to detect biases in the methods. 

Interpretability has been approached primarily from two main angles: \emph{model interpretability}, i.e., making the architecture itself interpretable, and \emph{prediction interpretability}, i.e., explaining particular predictions of the model (cf. \cite{Lei2016Rationalizing}). Requiring the model itself to be transparent is often too restrictive and challenging to achieve. Indeed, prediction interpretability can be more easily sought \emph{a posteriori} for black-box systems including neural networks. 

In this work, we propose a novel approach to prediction interpretability with only oracle access to the model generating the prediction. Following \cite{Ribeiro2016Why}, we turn the local behavior of the model around the given input into an interpretable representation of its operation. In contrast to previous approaches, we consider structured prediction where both inputs and outputs are combinatorial objects, and our explanation consists of a summary of operation rather than a simpler prediction method. 

Our method returns an ``explanation'' consisting of sets of input and output tokens that are causally related under the black-box model. Causal dependencies arise from analyzing perturbed versions of inputs that are passed through the black-box model. Although such perturbations might be available in limited cases, we generate them automatically. For sentences, we adopt a variational autoencoder to produce semantically related sentence variations. The resulting inferred causal dependencies (interval estimates) form a dense bi-partite graph over tokens from which explanations can be derived as robust min-cut k-partitions.

We demonstrate quantitatively that our method can recover known dependencies. As a starting point, we show that a grapheme-to-phoneme dictionary can be largely recovered if given to the method as a black-box model. We then show that the explanations provided by our method closely resemble the attention scores used by a neural machine translation system. Moreover, we illustrate how our summaries can be used to gain insights and detect biases in translation systems. Our main contributions are:
\begin{itemize}
	\vspace{-0.4cm}
	\item We propose a general framework for explaining structured black-box models
	\item For sequential data, we propose a variational autoencoder for controlled generation of input perturbations required for causal analysis
	\item We evaluate the explanations produced by our framework on various sequence-to-sequence prediction tasks, showing they can recover known associations and provide insights into the workings of complex systems.
\end{itemize}

\section{Related Work}

There is a wide body of work spanning various fields centered around the notion of ``interpretability''. This term, however, is underdetermined, so the goals, methods and formalisms of these approaches are often non-overlapping \cite{Lipton2016Mythos}. In the context of machine learning, perhaps the most visible line of work on interpretability focuses on medical applications \cite{Caruana2015Intelligible}, where trust can be a decisive factor on whether a model is used or not. With the ever-growing success and popularity of deep learning methods for image processing, recent work has addressed interpretability in this setting, usually requiring access to the method's activations and gradients \cite{selvaraju2016grad}, or directly modeling how influence propagates \cite{Bach2015Pixel-wise}. For a broad overview of interpretability in machine learning, we refer the reader to the recent survey by \citet{Doshi-Velez2017Roadmap}. 

Most similar to this work are the approaches of \citet{Lei2016Rationalizing} and \citet{Ribeiro2016Why}. The former proposes a model that justifies its predictions in terms of fragments of the input. This approach formulates explanation generation as part of the learning problem, and, as most previous work, only deals with the case where predictions are scalar or categorical. On the other hand, \citet{Ribeiro2016Why} propose a framework for explaining the predictions of black-box classifiers by means of locally-faithful interpretable models. They focus on sparse linear models as explanations, and rely on local perturbations of the instance to explain. Their model assumes the input directly admits a fixed size interpretable representation in euclidean space, so their framework operates directly on this vector-valued representation.

Our method differs from---and can be thought of as generalizing---these approaches in two fundamental aspects. First, our framework considers both inputs and outputs to be structured objects thus extending beyond the classification setting. This requires rethinking the notion of explanation to adapt it to variable-size combinatorial objects. Second, while our approach shares the locality and model-agnostic view of \citet{Ribeiro2016Why}, generating perturbed versions of structured objects is a challenging task by itself. We propose a solution to this problem in the case of sequence-to-sequence learning. 

\section{Interpreting structured prediction}

Explaining predictions in the structured input-structured output setting poses various challenges. As opposed to scalar or categorical prediction,  structured predictions vary in size and complexity. Thus, one must decide not only how to explain the prediction, but also what parts of it to explain. Intuitively, the ``size'' of an explanation should grow with the size of the input and output. A good explanation would ideally also decompose into \emph{cognitive chunks} \cite{Doshi-Velez2017Roadmap}: basic units of explanation which are a priori bounded in size. Thus, we seek a framework that naturally decomposes an explanation into (potentially several) \emph{explaining components}, each of which justifies, from the perspective of the black-box model, parts of the output relative to the parts of the input. 

Formally, suppose we have a black-box model $F: \mathcal{X} \rightarrow \mathcal{Y}$ that maps a structured input $\mathbf{x} \in \cX$ to a structured output $\mathbf{y} \in \cY$. We make no assumptions on the spaces $\cX,\cY$, except that their elements admit a feature-set representation $\mathbf{x} = \{x_1,x_2,\dots,x_n\}$, $\mathbf{y} = \{y_1, y_2, \dots, y_m\}$. Thus, $\mathbf{x}$ and $\mathbf{y}$ can be sequences, graphs or images. We refer to the elements $x_i$ and $y_j$ as units or ``tokens'' due to our motivating application of sentences, though everything in this work holds for other combinatorial objects.

For a given input output pair $(\mathbf{x},\mathbf{y})$, we are interested in obtaining an \emph{explanation} of $\mathbf{y}$ in terms of $\mathbf{x}$. Following \cite{Ribeiro2016Why}, we seek explanations via \emph{interpretable representations} that are both i) \emph{locally faithful}, in the sense that they approximate how the model behaves in the vicinity of $\mathbf{x}$, and ii) \emph{model agnostic}, that is, that do not require any knowledge of $F$. For example, we would like to identify whether token $x_i$ is a likely cause for the occurrence of $y_j$ in the output when the input context is $\mathbf{x}$. 
Our assumption is that we can summarize the behavior of $F$ around $\mathbf{x}$ in terms of a weighted bipartite graph $G = (V_x \cup V_y, E)$, where the nodes $V_x$ and $V_y$ correspond to the elements in $\mathbf{x}$ and $\mathbf{y}$, respectively, and the weight of each edge $E_{ij}$ corresponds to the influence of the occurrence of token $x_i$ on the appearance of $y_j$. The bipartite graph representation suggests naturally that the explanation be given in terms of explaining components. We can formalize these components as subgraphs $G^k = (V^k_x\cup V^k_y,E^k)$, where the elements in $V^k_x$ are likely causes for the elements in $V^k_y$. Thus, we define an explanation of $\mathbf{y}$ as a collection of such components: ${E}_{x \rightarrow y} = \{ G^1,\dots,G^k\}$.

\begin{figure*}
	\centering
	\includegraphics[scale=0.7, trim = {0 0.8cm 0cm 0cm}, clip]{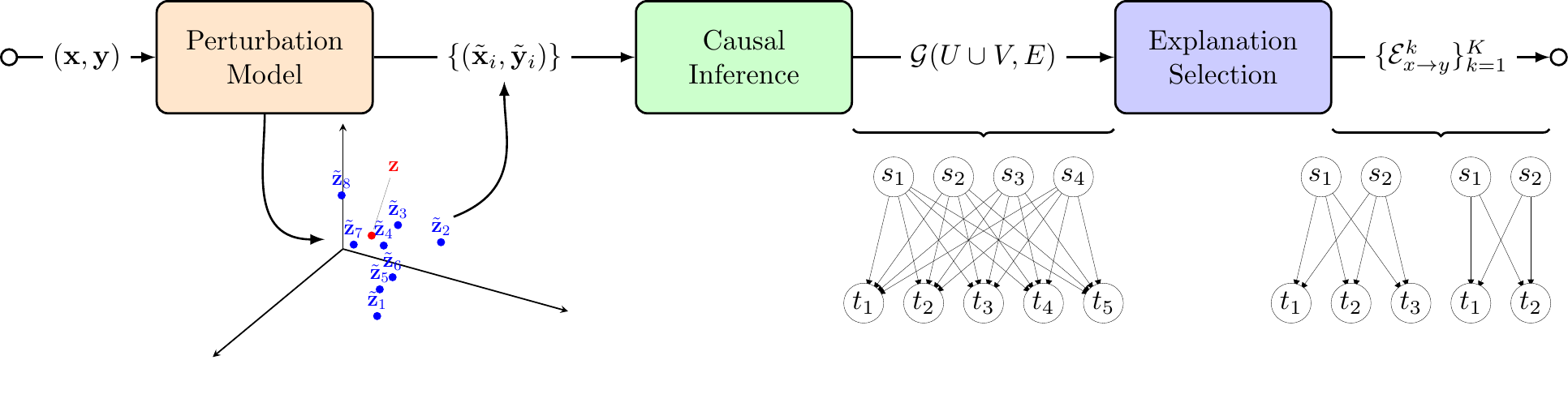}
	\caption{A schematic representation of the proposed prediction interpretability method.}\label{fig:GlobalDiagram}
\end{figure*}

Our approach formalizes this framework through a pipeline (sketched in Figure~\ref{fig:GlobalDiagram}) consisting of three main components, described in detail in the following section: a perturbation model for exercising $F$ locally, a causal inference model for inferring associations between inputs and predictions, and a selection step for partitioning and selecting the most relevant sets of associations. We refer to this framework as a \emph{structured-output causal rationalizer} (\textproc{SocRat}). 

\paragraph{A note on alignment models} 
\label{par:a_note_on_alignment_models}
When the inputs and outputs are sequences such as sentences, one might envision using an alignment model, such as those used in MT, to provide an explanation. This differs from our approach in several respects. Specifically, we focus on explaining the behavior of the ``black box'' mapping $F$ only locally, around the current input context, not globally. Any global alignment model would require access to substantial parallel data to train and would have varying coverage of the local context around the specific example of interest. Any global model would likely also suffer from misspecification in relation to $F$. A more related approach to ours would be an alignment model trained locally based on the same perturbed sentences and associated outputs that we generate. 

\section{Building blocks}

\subsection{Perturbation Model}
\label{sec:perturbation}

The first step in our approach consists of obtaining \emph{perturbed} versions of the input: semantically similar to the original but with potential changes in elements and their order. This is a major challenge with any structured inputs. We propose to do this using a variational autoencoder (VAE) \citep{Kingma2014Autoencoding, Rezende2014Stochastic}. VAEs have been successfully used with fixed dimensional inputs such as images \cite{Rezende2015Variational, Sonderby2016Ladder} and recently also adapted to generating sentences from continuous representations \cite{Bowman2016Generating}. The goal is to introduce the perturbation in the continuous latent representation rather than directly on the structured inputs. 

A VAE is composed of a probabilistic encoder $\textsc{Enc}: \cX \rightarrow \R^d$ and a decoder $\textsc{Dec}: \R^d\rightarrow \cX$. The encoder defines a distribution over latent codes $q(\z|\x)$, typically by means of a two-step procedure that first maps $\x \mapsto (\bm{\mu}, \bm{\sigma})$ and then samples $\z$ from a gaussian distribution with these parameters. We can leverage this stochasticity to obtain perturbed versions of the input by sampling repeatedly from this distribution, and then mapping these back to the original space using the decoder. The training regime for the VAE ensures approximately that a small perturbation of the hidden representation maintains similar semantic content while introducing small changes in the decoded surface form. We emphasize that the approach would likely fail with an ordinary autoencoder where small changes in the latent representation can result in large changes in the decoded output.  In practice, we ensure diversity of perturbations by scaling the variance term $\bm{\sigma}$ and sampling points $\tilde{\z}$ and different resolutions. We provide further details of this procedure in the Appendix. Naturally, we can train this perturbation model in advance on (unlabeled) data from the input domain $\cX$, and then use it as a subroutine in our method. After this process is complete, we have $N$ pairs of perturbed input-output pairs: $\{(\tilde{\mathbf{x}}_i, \tilde{\mathbf{y}}_i)\}_{i=1}^N$ which exercise the mapping $F$ around semantically similar inputs.

\subsection{Causal model}
\label{sec:causal}

The second step consists of using the perturbed input-output pairs 
$\{(\tilde{\mathbf{x}}_i, \tilde{\mathbf{y}}_i)\}_{i=1}^N$ to infer causal dependencies between the original input and output tokens. A naive approach would consider 2x2 contingency tables representing presence/absence of input/output tokens together with a test statistic for assessing their dependence. Instead, we incorporate \emph{all} input tokens simultaneously to predict the occurrence of a single output token via logistic regression. The quality of these dependency estimators will depend on the frequency with which each input and output token occurs in the perturbations. Thus, we are interested in obtaining uncertainty estimates for these predictions, which can be naturally done with a Bayesian approach to logistic regression. Let $\phi_{\mathbf{x}}(\tilde{\mathbf{x}}) \in \{0,1\}^{|\mathbf{x}|}$ be a binary vector encoding the presence of the original tokens $x_1,\dots,x_n$ from $\mathbf{x}$ in the perturbed version $\tilde{\mathbf{x}}$. For each target token $y_j\in \mathbf{y}$, we estimate a model:
\begin{equation}\label{eq:logreg}
 		P(y_j \in \tilde{\mathbf{y}} \st \tilde{\mathbf{x}}) = 
		\sigma(\boldsymbol{\theta}^T_j \phi_{\mathbf{x}}(\mathbf{\tilde{x}}) )
\end{equation}
where $\sigma(z) = (1+\exp(-z))^{-1}$. We use a Gaussian approximation for the logarithm of the logistic function together with the prior $p(\bm{\theta})=\mathcal{N}(\bm{\theta}_0,\mathbf{H}_0^{-1})$ \cite{Murphy2012Machine}. Since in our case all tokens are guaranteed to occur at least once (we include the original example pair as part of the set), we use $\bm{\theta}_0 = \alpha \mathbf{1}, \mathbf{H}_0 = \beta \mathbf{I}$, with $\alpha,\beta >0$.  Upon completion of this step, we have dependency coefficients between all original input and output tokens $\{\theta_{ij}\}$, along with their uncertainty estimates.

\subsection{Explanation Selection}\label{sec:selection}

The last step in our interpretability framework consists of selecting a set explanations for $(\mathbf{x},\mathbf{y})$. The steps so far yield a dense bipartite graph between the input and output tokens. Unless $|\mathbf{x}|$ and $|\mathbf{y}|$ are small, this graph itself may not be sufficiently interpretable. We are interested in selecting \emph{relevant} components of this dependency graph, i.e., partition the vertex set of $\mathcal{G}$ into disjoint subsets so as to minimize the weight of omitted edges (i.e.~the k-cut value of the partition).

\emph{Graph partitioning} is a well studied NP-complete problem \cite{Garey1976Simplified}. The usual setting assumes deterministic edge weights, but in our case we are interested in incorporating the uncertainty of the dependency estimates---resulting from their  finite sample estimation---into the partitioning problem. For this, we rely on the approach of \citet{Fan2012Robust} designed for interval estimates of edge weights. At a high level, this is a robust optimization formulation which seeks to minimize worst case cut values, and can be cast as a Mixed Integer Programming (MIP) problem. Specifically, for a bipartite graph $G=(U,V,E)$ with edge weights given as uncertainty intervals $\theta_{ij}\pm \hat{\theta}_{ij}$, the partitioning problem is given by
\begin{multline}\label{uncert_obj}
	\min_{(x_{ik}^u, x_{jk}^v, y_{ij}) \in Y} \sum_{i=1}^n \sum_{j=1}^m \theta_{ij}y_{ij}  +  \\
	\max_{ \substack{ S: S \subseteq J, |S| \leq \Gamma \\ (i_t, j_t) \in J \setminus S }} \sum_{(i,j) \in S} \hat{\theta}_{ij}y_{ij} + (\Gamma - \floor{\Gamma})\hat{\theta}_{i_t,j_t}y_{i_t,j_t}
\end{multline}
where $J = \{ (i,j) \st \theta_{ij} >0\}$, $x^u_{ik}$, $x^v_{jk}$ are indicators for vertex sets $U$ and $V$ respectively, $y_{ij}$ are binary auxiliary variables indicating whether $i$ and $j$ are in different partitions, and $Y$ is a set of constraints that ensure the K-partition is valid. $\Gamma$ is a parameter in $[0, |V|]$ which adjusts the robustness of the partition (the number of deviations from the mean edge values). Detailed explanation of this objective is provided in the Appendix.

\begin{algorithm}[t]
  \caption{Structured-output causal rationalizer}\label{algo:socrat}
  \begin{algorithmic}[1]
    \Procedure{Socrat}{$\mathbf{x},\mathbf{y},F$}
  \State $(\bm{\mu},\bm{\sigma}) \gets \Call{Encode}{\mathbf{x}}$
	\For{$i=1$ {\bfseries to} $N$}
    \State $\tilde{\mathbf{z}}_i \gets \Call{Sample}{\bm{\mu},\bm{\sigma}}$\hspace*{2em}%
        \rlap{\smash{$\left.\begin{array}{@{}c@{}}\\{}\\{}\\{}\\{}\end{array}\color{black}\right\}%
          \color{black}\begin{tabular}{l}Perturbation \\ Model.\end{tabular}$}}
    \State $\tilde{\mathbf{x}}_i \gets \Call{Decode}{\tilde{\mathbf{z}}_i}$
    \State $\tilde{\mathbf{y}}_i \gets F(\tilde{\mathbf{x}}_i)$
    \EndFor
   \State $G \hspace{0.6cm} \gets \Call{Causal}{\mathbf{x},\mathbf{y},\{\tilde{\mathbf{x}}_i, \tilde{\mathbf{y}}_i\}_{i=1}^N}$
   \State $E_{x \mapsto y}\gets \Call{Bipartition}{G}$
   \State $E_{x \mapsto y}\gets \Call{sort}{{E}_{x \mapsto y}}$\Comment{{\footnotesize By cut capacity}}
    \State \textbf{return} ${E}_{x \mapsto y}$
  \EndProcedure
\end{algorithmic}
\end{algorithm}

If $|\mathbf{x}|$ and $|\mathbf{y}|$ are small, the number of clusters $K$ will also be small, so we can simply return all the partitions (i.e.~the \emph{explanation chunks}) $E_{x\rightarrow y}^k:= (V^k_x \cup V^k_y)$. However, when $K$ is large, one might wish to entertain only the $\kappa$ most relevant explanations. The graph partitioning framework provides us with a natural way to score the importance of each chunk. Intuitively, subgraphs that have few high-valued edges connecting them to other parts of the graph (i.e.~low \emph{cut-capacity}) can be thought of as \emph{self-contained} explanations, and thus more relevant for interpretability. We can therefore define the importance score an atom as:
\begin{align}\label{eq:partcut}
	\text{importance}(E_{x \rightarrow y}^k) &:= 
    -	\sum_{(i,j) \in X_k} \theta_{ij}
\end{align}
where $X_k$ is the cut-set implied by $E_{x \rightarrow y}^k$:
\[ X_k = \{ (i,j) \in E \st i \in E_{x \rightarrow y}^k, \medspace j \in V \setminus E_{x \rightarrow y}^k \} \]
The full interpretability method is succinctly expressed in Algorithm \ref{algo:socrat}.

\section{Experimental Framework}

\subsection{Training and optimization}

For the experiments involving sentence inputs, we train in advance the VAE described in Section~\ref{sec:perturbation}. We use symmetric encoder-decoders consisting of recurrent neural networks with an intermediate variational layer. In our case, however, we use $L$ stacked RNN's on both sides, and a stacked variational layer. Training variational autoencoders for text is notoriously hard. In addition to dropout and KLD annealing \cite{Bowman2016Generating}, we found that slowly scaling the variance sampled from the normal distribution from $0$ to $1$ made training much more stable. 

For the partitioning step we compare the robust formulation described above with two classical approaches to bipartite graph partitioning which do not take uncertainty into account: the coclustering method of \citet{Dhillon2001Coclustering} and the biclustering method of \citet{Kluger2003Spectral}. For these two, we use off-the-shelf implementations,\footnote{http://scikit-learn.org/stable/modules/biclustering.html} while we solve the MIP problem version of \eqref{uncert_obj} with the optimization library \texttt{gurobi}.\footnote{http://www.gurobi.com/}


\subsection{Recovering simple mappings} 
\label{sub:recovering_known_mappings}

Before using our interpretability framework in real tasks where quantitative evaluation of explanations is challenging, we test it in a simplified setting where the ``black-box'' is simple and fully known. A reasonable minimum expectation on our method is that it should be able to infer many of these simple dependencies. For this purpose, we use the CMU Dictionary of word pronunciations,\footnote{www.speech.cs.cmu.edu/cgi-bin/cmudict} which is based on the ARPAbet symbol set and consists of about 130K word-to-phoneme pairs. Phonemes are expressed as tokens of 1 to 3 characters. An example entry in this dictionary is the pair $\textit{vowels} \mapsto \texttt{V AW1 AH0 L Z}$. Though the mapping is simple, it is not one-to-one (a group of characters can correspond to a single phoneme) nor deterministic (the same character can map to different phonemes depending on the context). Thus, it provides a reasonable testbed for our method.  The setting is as follows: given an input-output pair from the \texttt{cmudict} ``black-box'', we use our method to infer dependencies between characters in the input and phonemes in the output. Since locality in this context is morphological instead of semantic, we produce perturbations selecting $n$ words randomly from the intersection of the \texttt{cmudict} vocabulary and the set of words with edit distance at most 2 from the original word.

\begin{figure}[t]
	\centering
	\includegraphics[scale=0.31]{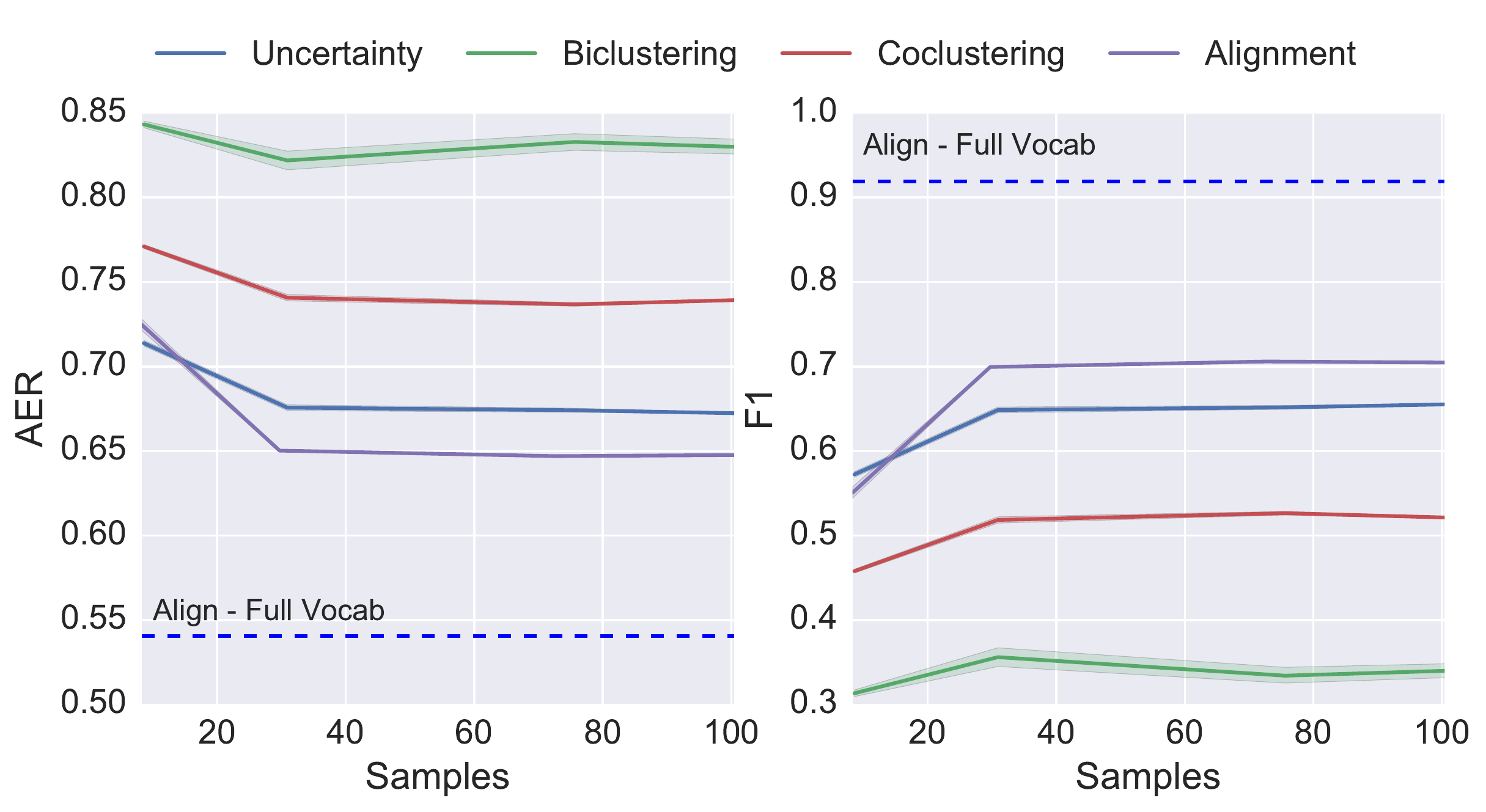}
	\caption{Arpabet test results as a function of number of perturbations used. Shown are mean plus confidence bounds over 5 repetitions. \textbf{Left}: Alignment Error Rate, \textbf{Right}: F1 over edge prediction.}\label{fig:arpabet_AER}
\end{figure}

To evaluate the inferred dependencies, we randomly selected 100 key-value pairs from the dictionary and manually labeled them with character-to-phoneme alignments. Even though our framework is not geared to produce pairwise alignments, it should nevertheless be able to recover them to a certain extent. To provide a point of reference, we compare against a (strong) baseline that is tailored to such a task: a state-of-the-art unsupervised word alignment method based on Monte Carlo inference \cite{Tiedmann2016Efficient}. The results in Figure~\ref{fig:arpabet_AER} show that the version of our method that uses the uncertainty clustering performs remarkably close to the alignment system, with an alignment error rate only ten points above an oracle version of this system that was trained on the \emph{full} arpabet dictionary (dashed line). The raw and partitioned explanations provided by our method for an example input-output pair are shown in Table~\ref{fig:arpabet_bipartite}, where the edge widths correspond to the estimated strength of dependency. Throughout this work we display the nodes in the same lexical order of the inputs/outputs to facilitate reading, even if that makes the explanation chunks less visibly discernible. Instead, we sometimes provide an additional (sorted) heatplot of dependency values to show these partitions.

\begin{table}
	\setlength{\tabcolsep}{2pt}
	\small
	\begin{tabular}{p{3.25cm} p{0.15cm}  p{3.6cm}}
		\toprule
		\hspace{0.4cm} Raw Dependencies &  & \hspace{0.4cm} Explanation Graph \\
		\midrule
		\begin{minipage}{0.45\linewidth}
			\includegraphics[scale=0.25,trim={2.2cm 1.5cm 1.5cm 1cm},clip]{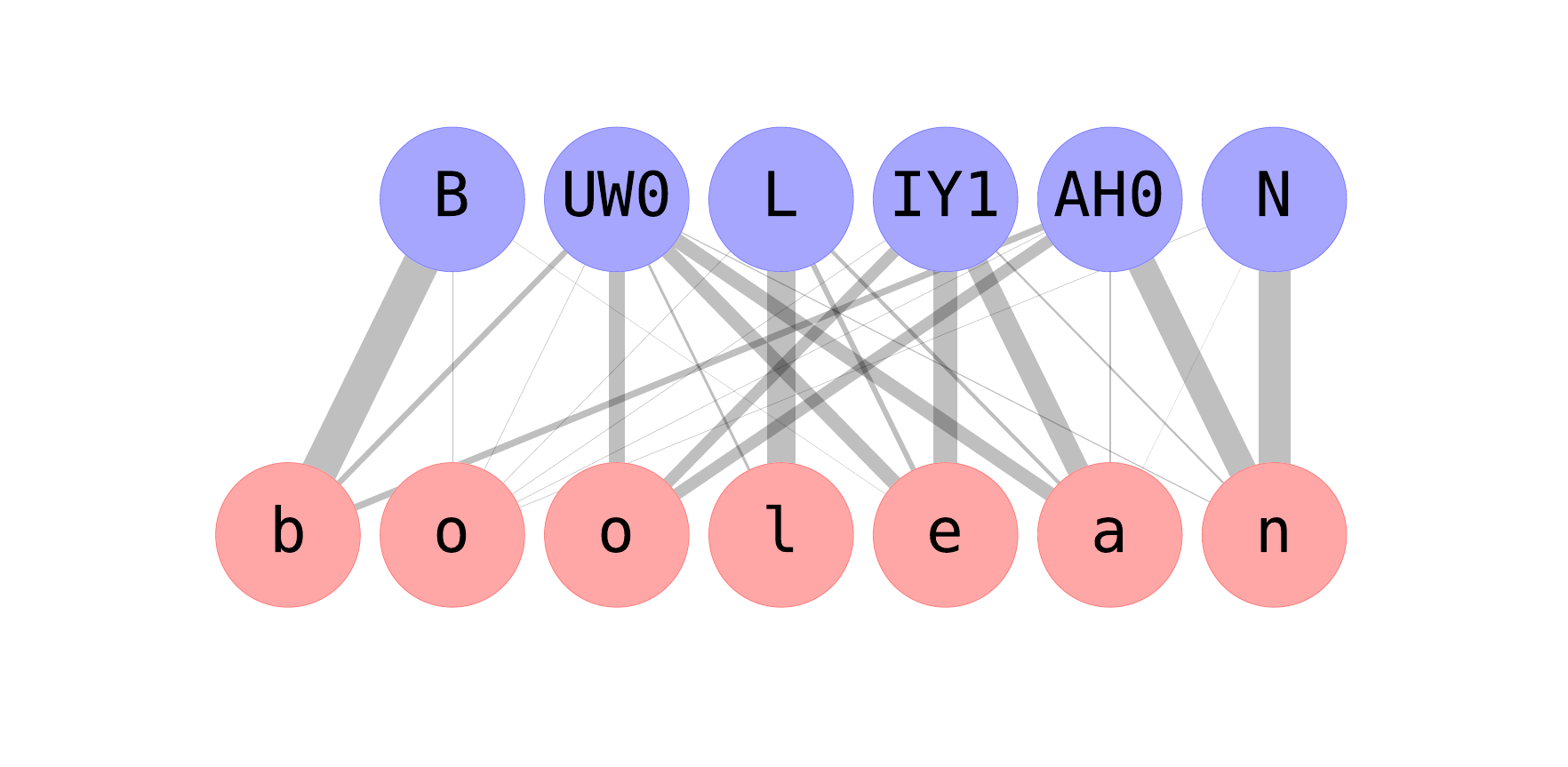}
		\end{minipage} 	&
		$\rightarrow$ &
		\begin{minipage}{0.45\linewidth}
			\includegraphics[scale=0.25,trim={2.2cm 1.5cm 1.5cm 1cm},clip]{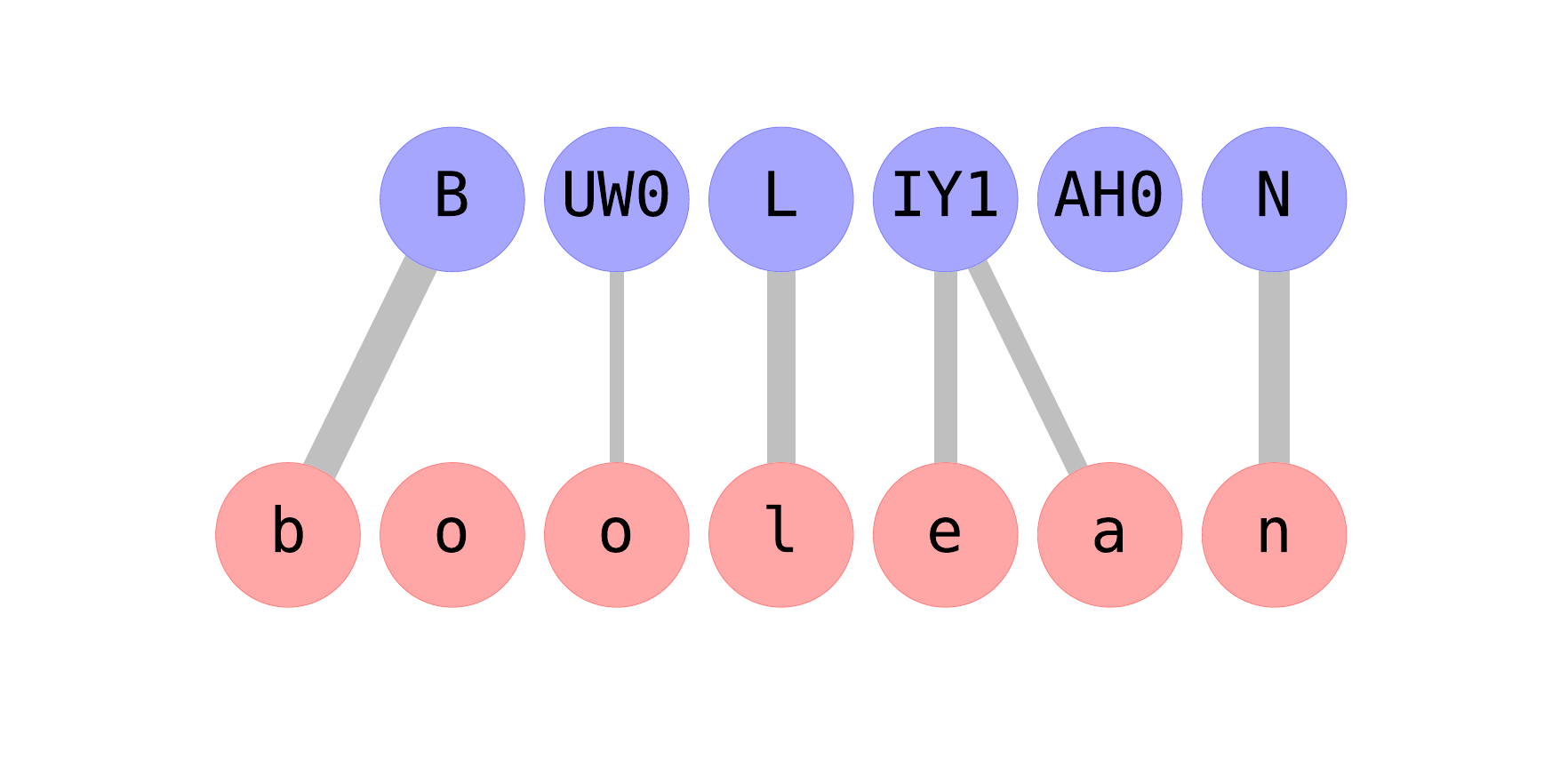}
		\end{minipage}
		\\
		\midrule
		\begin{minipage}{0.55\linewidth}
			\includegraphics[scale=0.25,trim={2.2cm 1.5cm 1.5cm 1cm},clip]{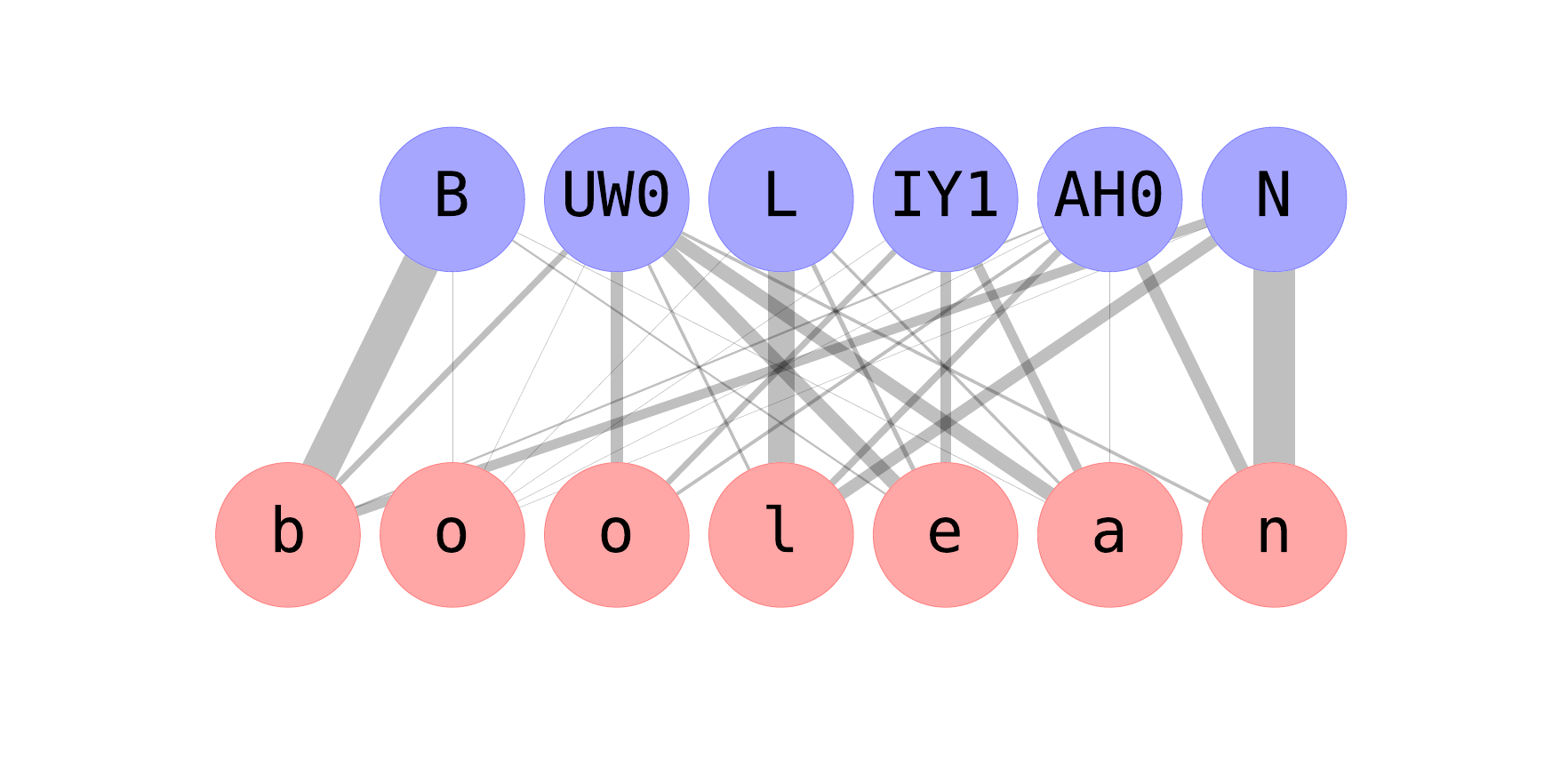}
		\end{minipage} &$\rightarrow$&
		\begin{minipage}{0.55\linewidth}
			\includegraphics[scale=0.25,trim={2.2cm 1.5cm 1.5cm 1cm},clip]{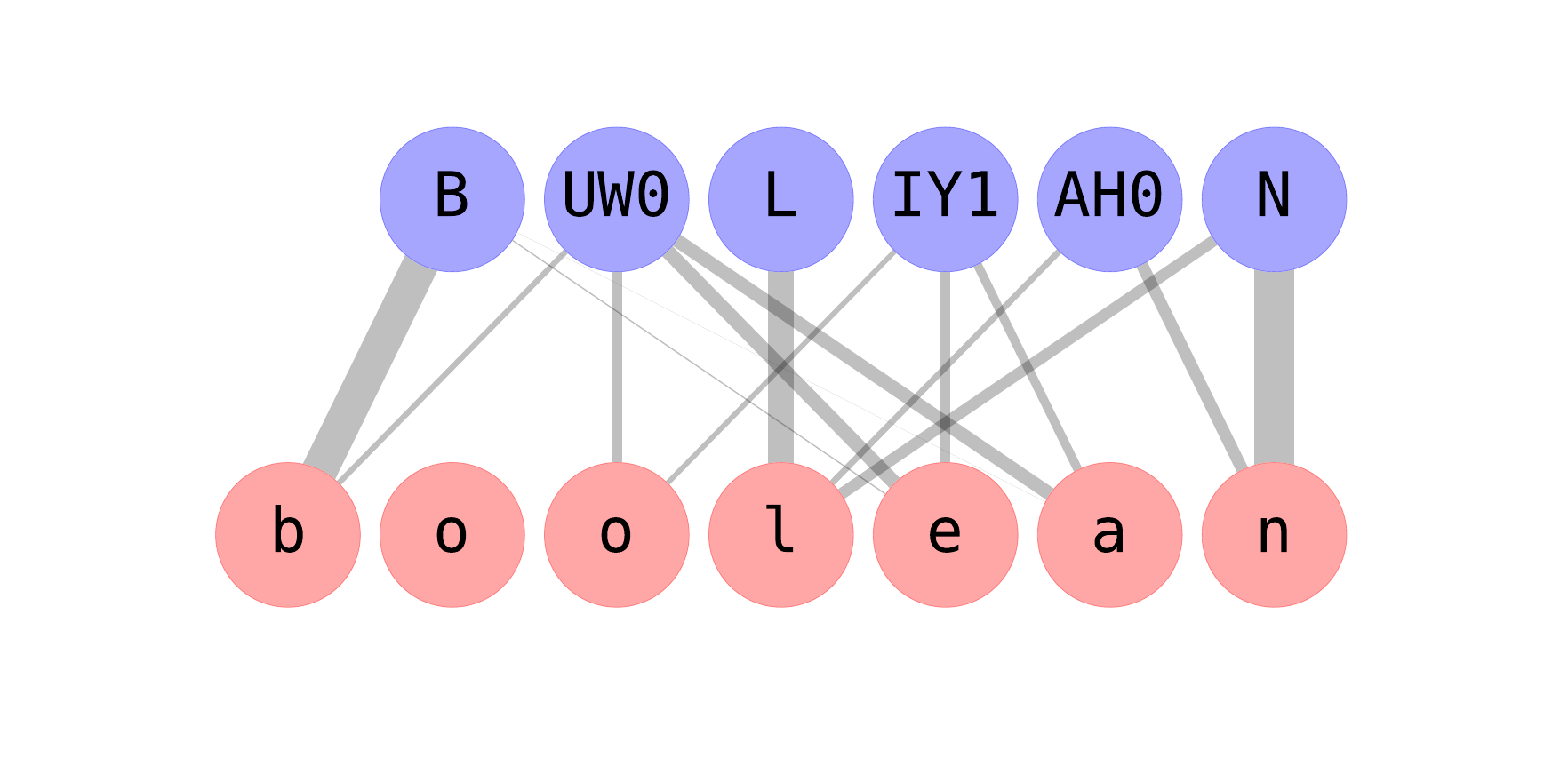}
		\end{minipage}
		\\
		\bottomrule
	\end{tabular}
	\caption{Inferred dependency graphs before (left) and after (right) explanation selection for the prediction: \emph{boolean} $\mapsto$ \texttt{B UW0 L IY1 AH0 N}, in independent runs with large (top) and small (bottom) clustering parameter $k$.}\label{fig:arpabet_bipartite}
\end{table}

\subsection{Machine Translation} 
\label{sub:machine_translation}

In our second set of experiments we evaluate our explanation model in a relevant and popular sequence-to-sequence task: machine translation. As black-boxes, we use three different methods for translating English into German: (i) Azure's Machine Translation system, (ii) a Neural MT model, and (iii) a human (native speaker of German). We provide details on all three systems in the Appendix. We translate the same English sentences with all three methods, and explain their predictions using \textsc{SocRat}. To be able to generate sentences with similar language and structure as those used to train the two automatic systems, we use the monolingual English side of the WMT14 dataset to train the variational autoencoder described in Section~\ref{sec:perturbation}. For every explanation instance, we sample $S=100$ perturbations and use the black-boxes to translate them. In all cases, we use the same default \textsc{SocRat} configurations, including the robust partitioning method.

\begin{figure}
	\includegraphics[scale=0.14]{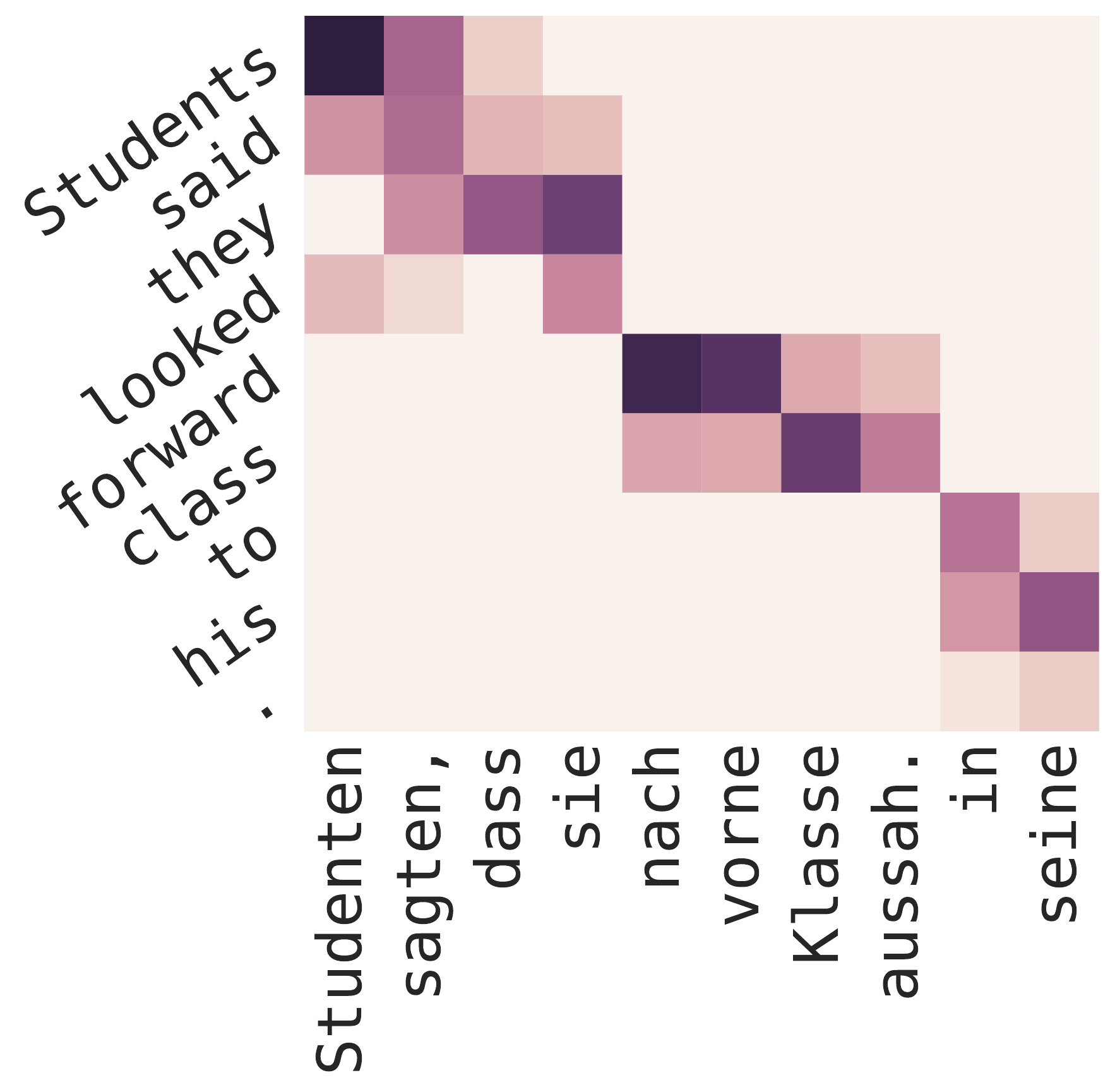}%
	\includegraphics[scale=0.3, trim={4cm 0cm 2.75cm 0.5cm}, clip]{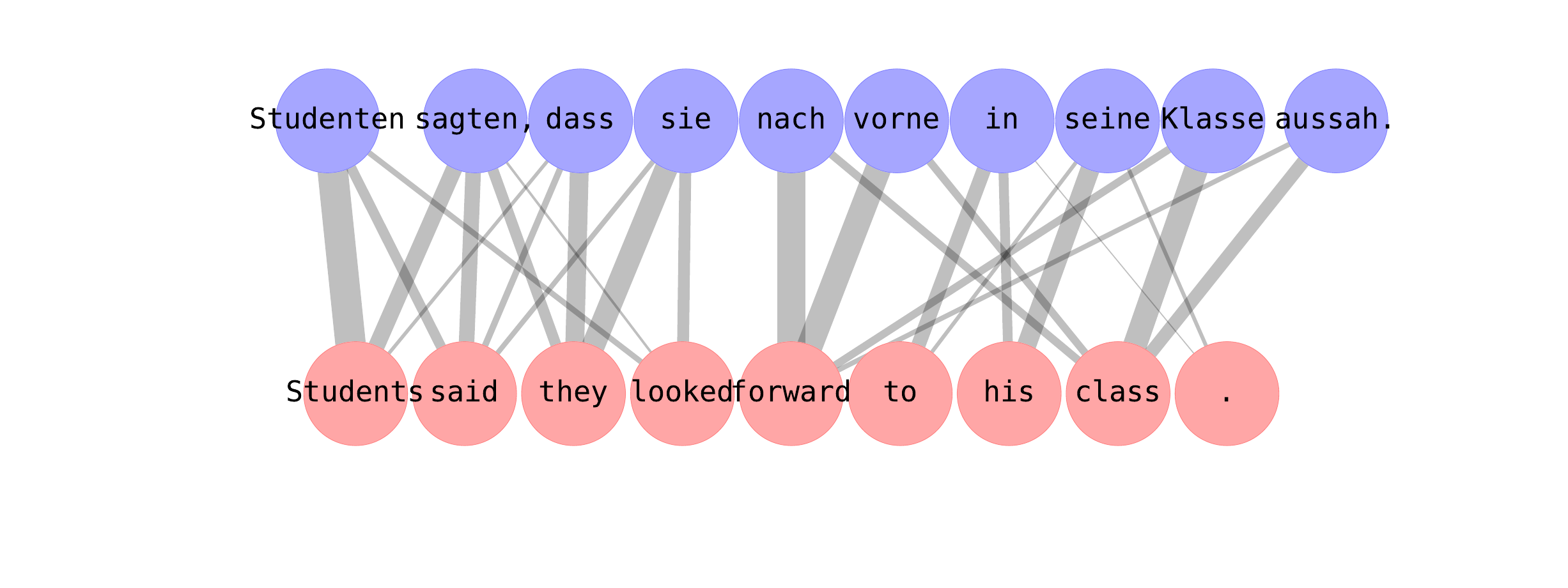}\\
	\includegraphics[scale=0.13]{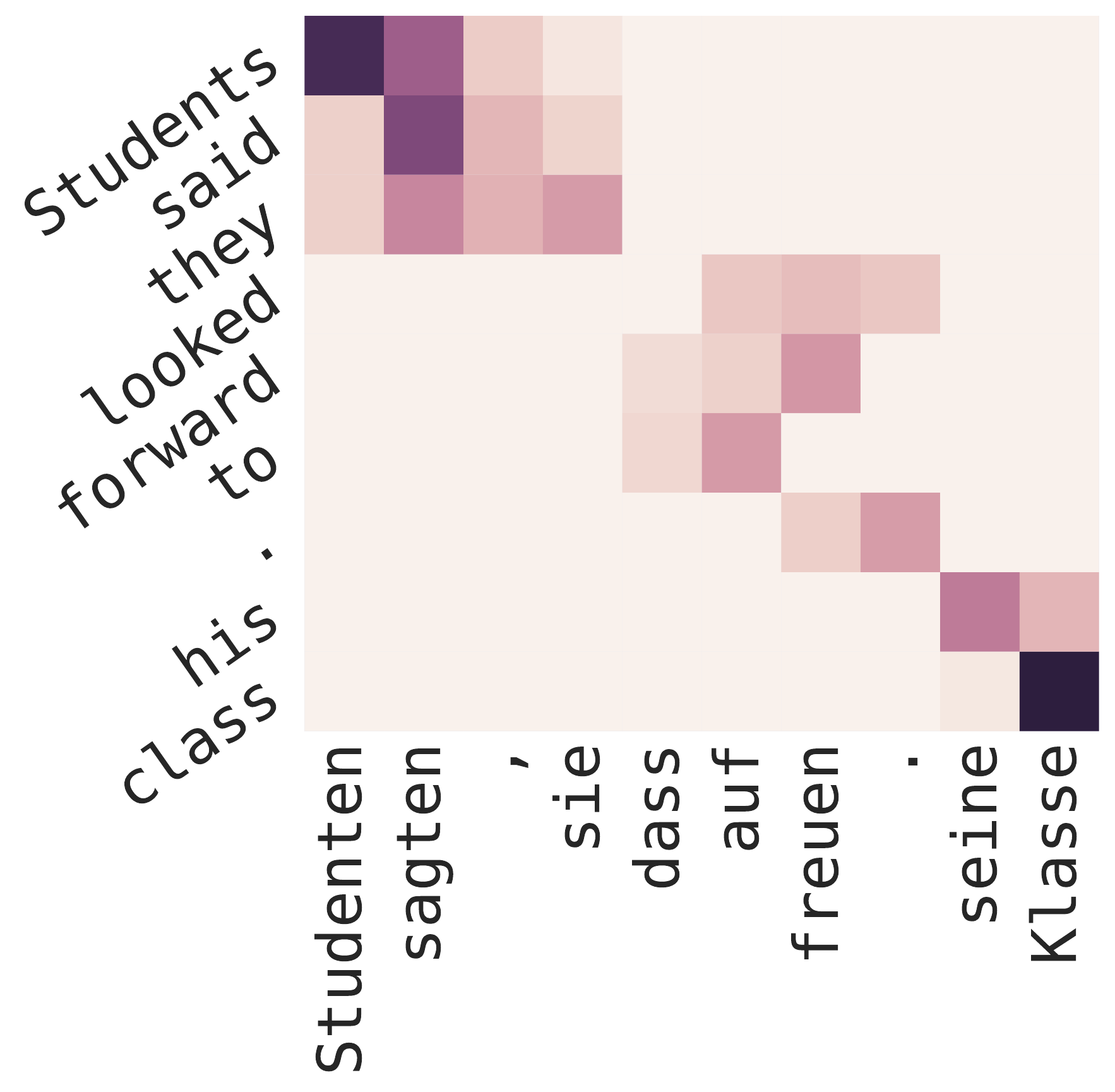}%
	\includegraphics[scale=0.3, trim={4cm 0.6cm 2.75cm 0.5cm}, clip]{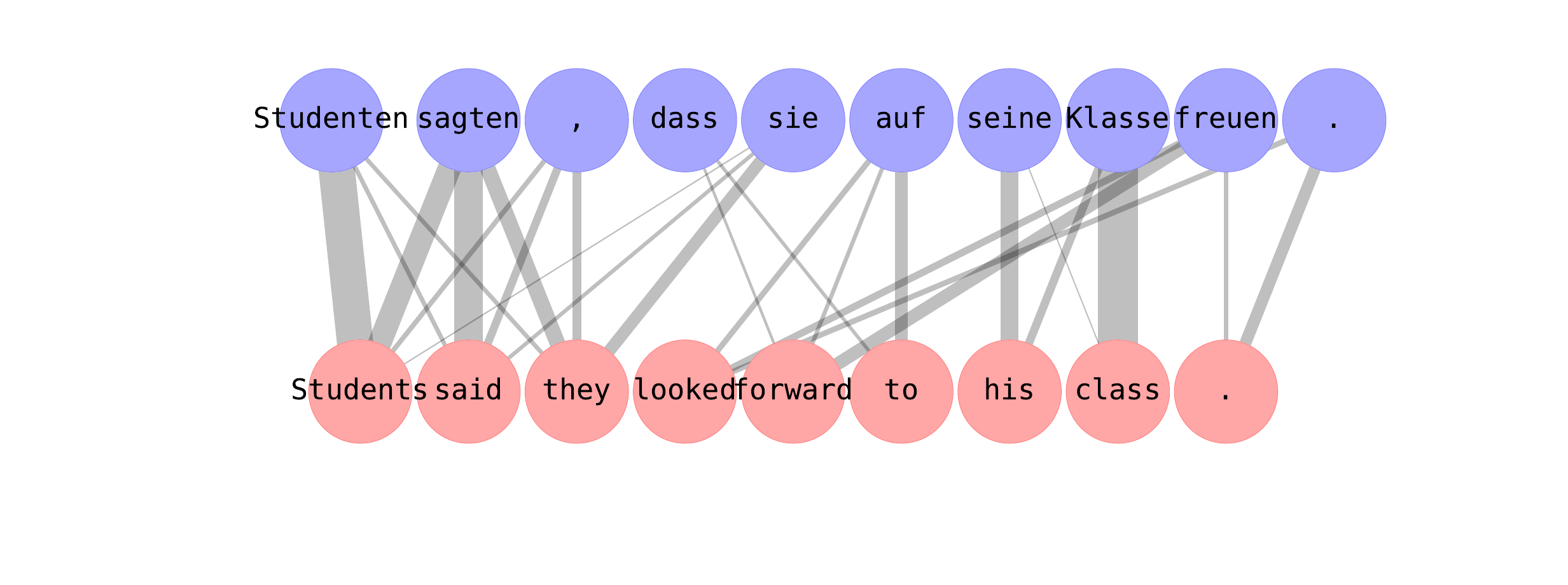}\\
	\includegraphics[scale=0.13]{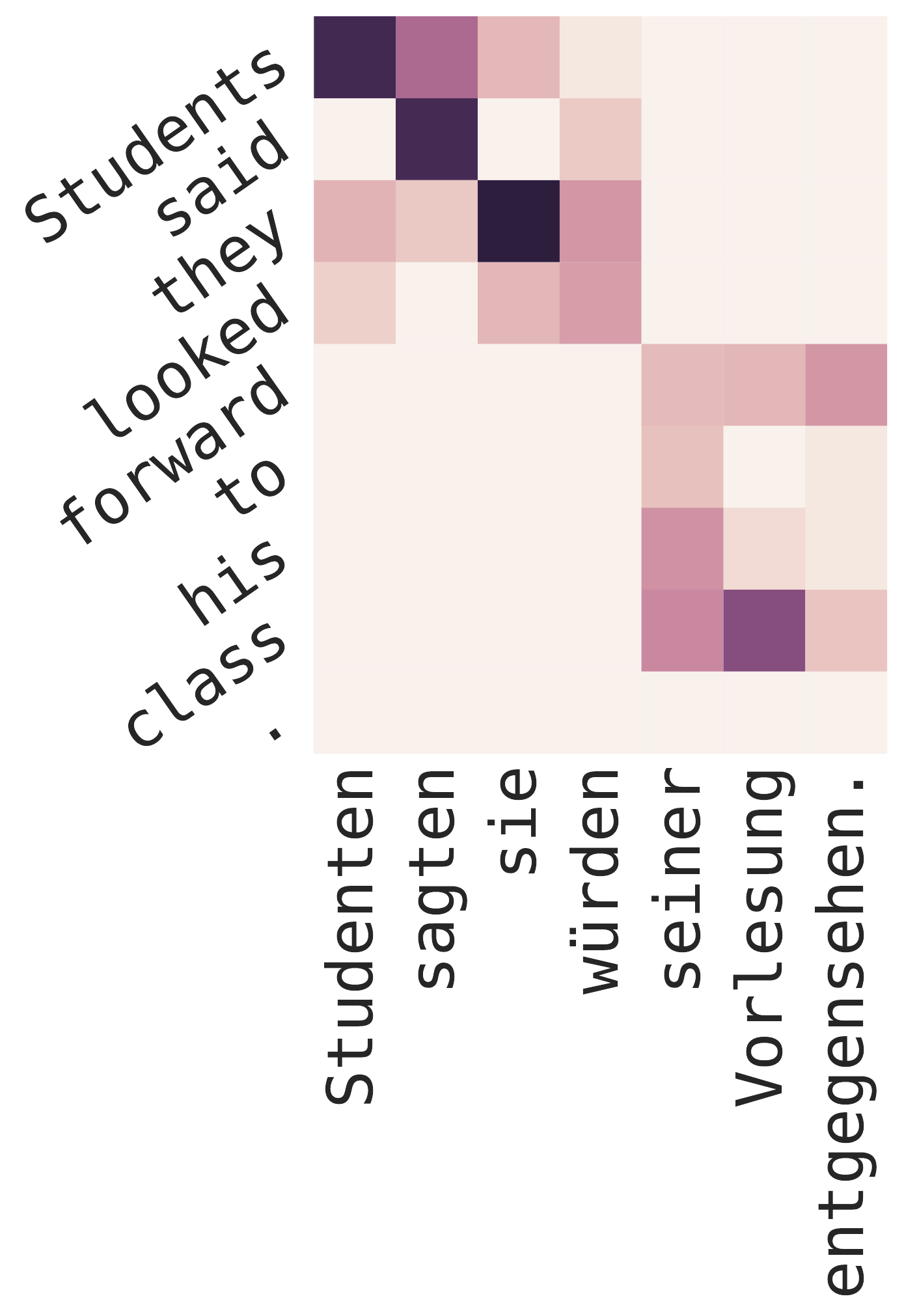}%
	\includegraphics[scale=0.3, trim={2cm 0cm 2.5cm 0.5cm}, clip]{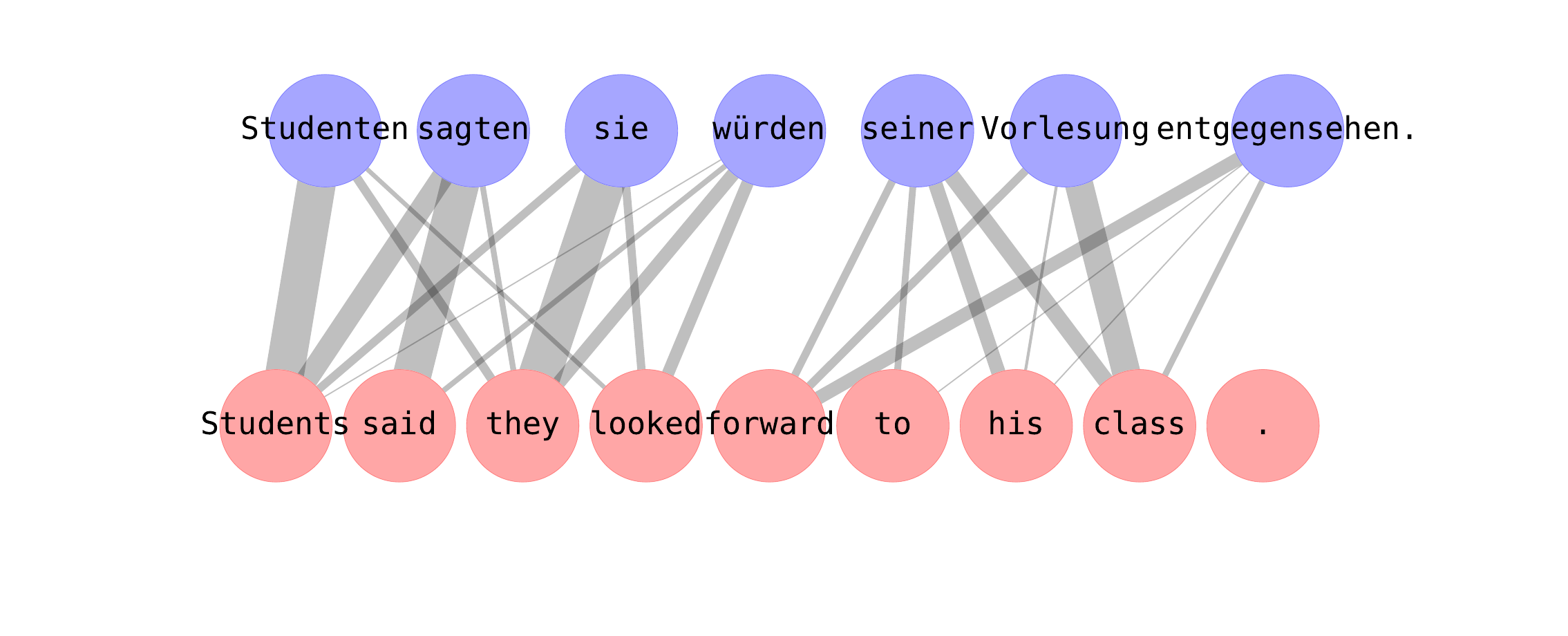}\\
	\vspace{-0.5cm}	
	\caption{Explanations for the predictions of three Black-Box translators: Azure (top), NMT (middle) and human (bottom). Note that the rows and columns of the heatmaps are permuted to show explanation \emph{chunks} (clusters).}\label{fig:explanations_mt}
\end{figure}

In Figure~\ref{fig:explanations_mt}, we show the explanations provided by our method for the predictions of each of the three systems on the input sentence \emph{``Students said they looked forward to his class''}. Although the three black-boxes all provided different translations, the explanations show a mostly consistent clustering around the two phrases in the sentence, and in all three cases the cluster with the highest cut value (i.e.~the most relevant explanative chunk) is the one containing the subject. Interestingly, the dependency coefficients are overall higher for the human than for the other systems, suggesting more coherence in the translations (potentially because the human translated sentences in context, while the two automatic systems carry over no information from one example to the next).

\begin{figure}
	\centering
	\includegraphics[scale=0.3]{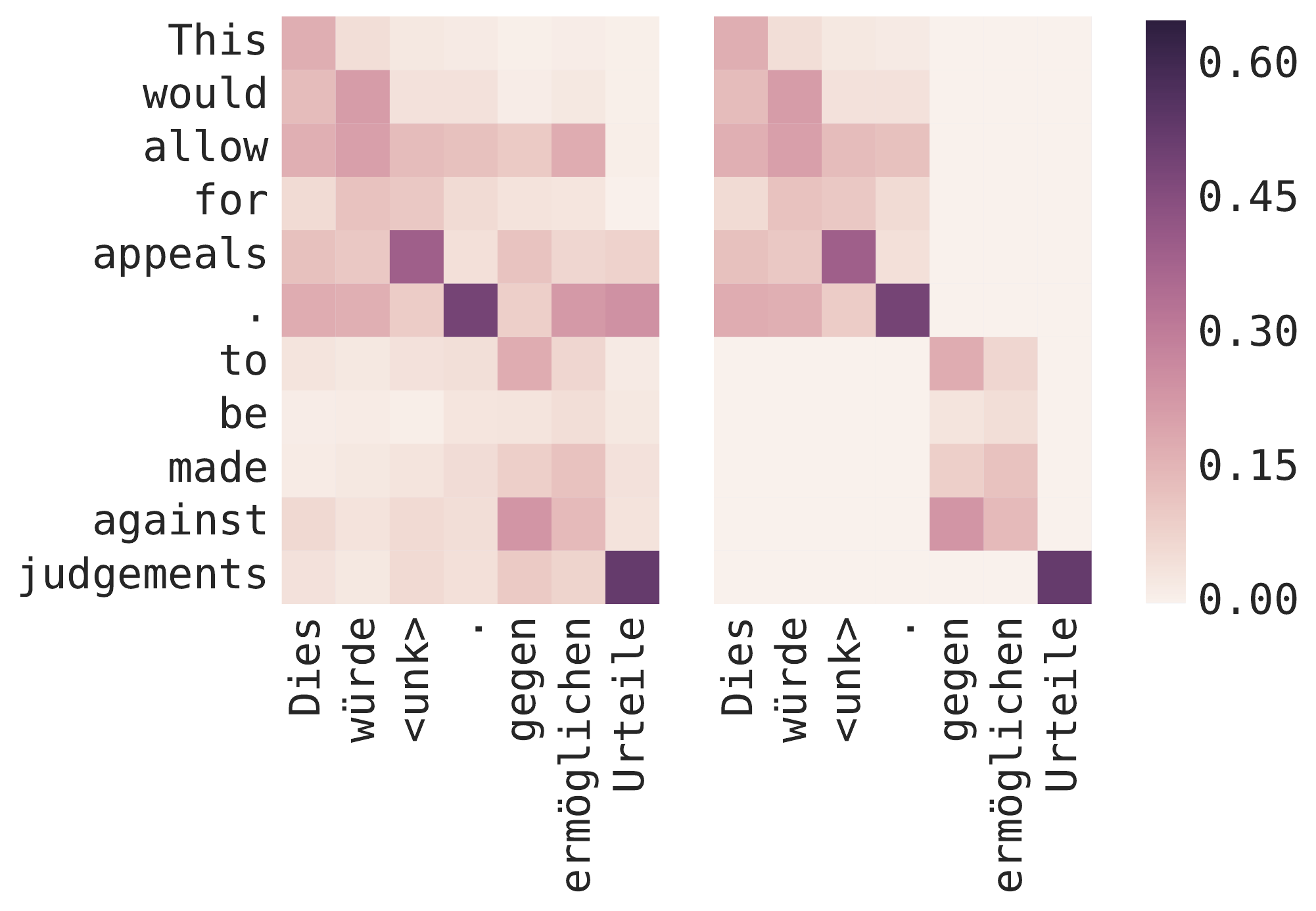}\\
	\includegraphics[scale=0.14]{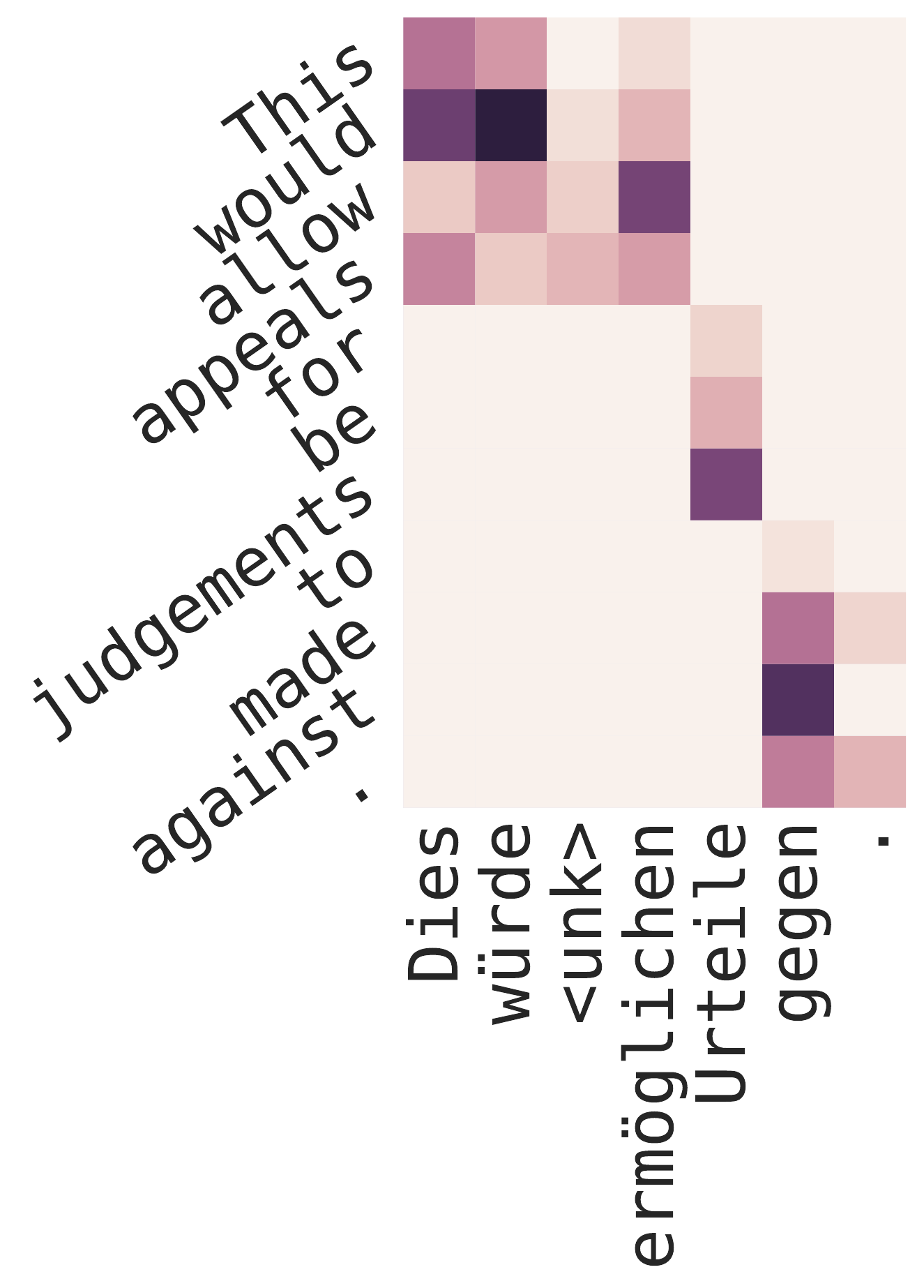}%
	\includegraphics[scale=0.28,trim={3.8cm 0cm 2.75cm 0.5cm},clip]{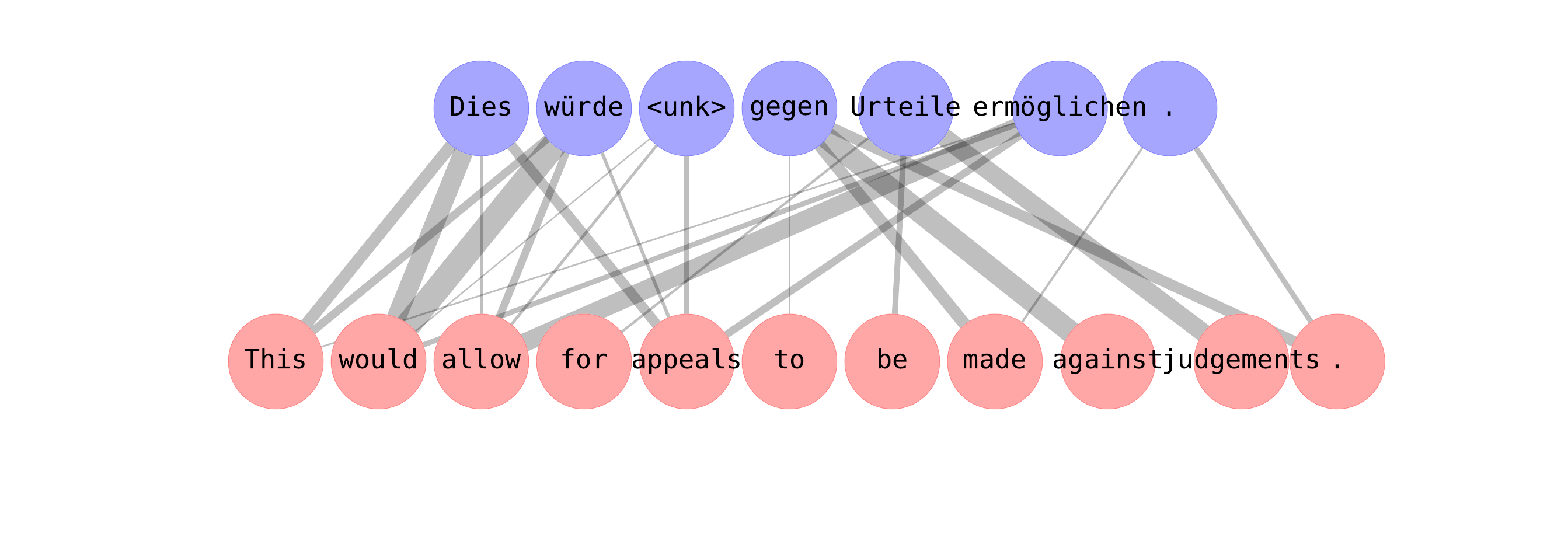}
	\caption{\textbf{Top}: Original and clustered attention matrix of the NMT system for a given translation. \textbf{Bottom}: Dependency estimates and explanation graph generated by \textsc{SocRat} with with $S=100$.}\label{fig:attn_vs_expl}
\end{figure}

The NMT system, as opposed to the other two, is not truly a black-box. We can \emph{open the box} to get a glimpse on the true dependencies on the inputs used by the system at prediction time (the attention weights) and compare them to the explanation graph. The attention matrix, however, is dense and not normalized over target tokens, so it is not directly comparable to our dependency scores. Nevertheless, we can partition it with the coclustering method described in Section~\ref{sec:selection} to enforce group structure and make it easier to compare. Figure~\ref{fig:attn_vs_expl} shows the attention matrix and the explanation for an example sentence of the test set. Their overall cluster structure agrees, though our method shows conservatism with respect to the dependencies of the function words (\emph{to}, \emph{for}). Interestingly, our method is able to figure out that the \texttt{<unk>} token was likely produced by the word ``appeals'', as shown by the explanation graph. 

It must be emphasized that although we display attention scores in various experiments in this work, we do so only for qualitative evaluation purposes. Our model-agnostic framework can be used on top of models that do not use attention mechanisms or for which this information is hard to extract. Even in cases where it is available, the explanation provided by \textsc{SocRat} might be complementary or even preferable to attention scores because: (a) being normalized on both directions (as opposed to only over source tokens) and partitioned, it is often more interpretable than a dense attention matrix, and (b) it can be retrieved \emph{chunk}-by-\emph{chunk} in decreasing order of relevance, which is especially important when explaining large inputs and/or outputs.

\subsection{A (mediocre) dialogue system} 
\label{sub:dialogue_systems}

So far we have used our method to explain (mostly) correct predictions of meaningful models. But we can use it to gain insights into the workings of flawed black-box systems too. To test this, we train a simple dialogue system on the OpenSubtitle corpus \cite{Tiedemann2009News}, consisting of $\sim$14M two-step movie dialogues. As before, we use a sequence-to-sequence model with attention, but now we constrain the quality of the model, using only two layers, hidden state dimension of 1000 and no hyper-parameter tuning.

\begin{figure}
	\centering
	\includegraphics[scale=0.35,trim={2.4cm 0cm 1.5cm 0.3cm},clip]{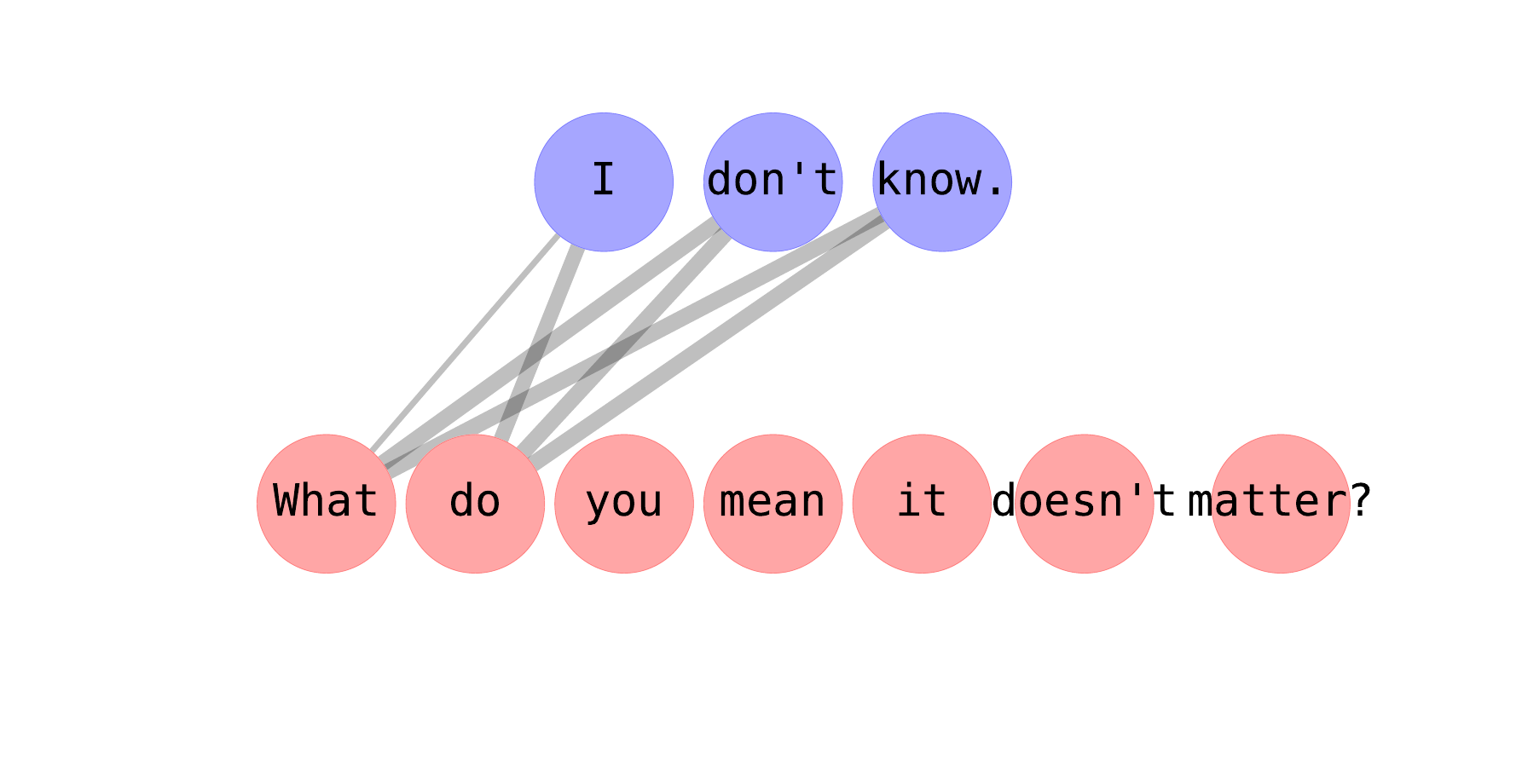}\hfill{}%
	\includegraphics[scale=0.23]{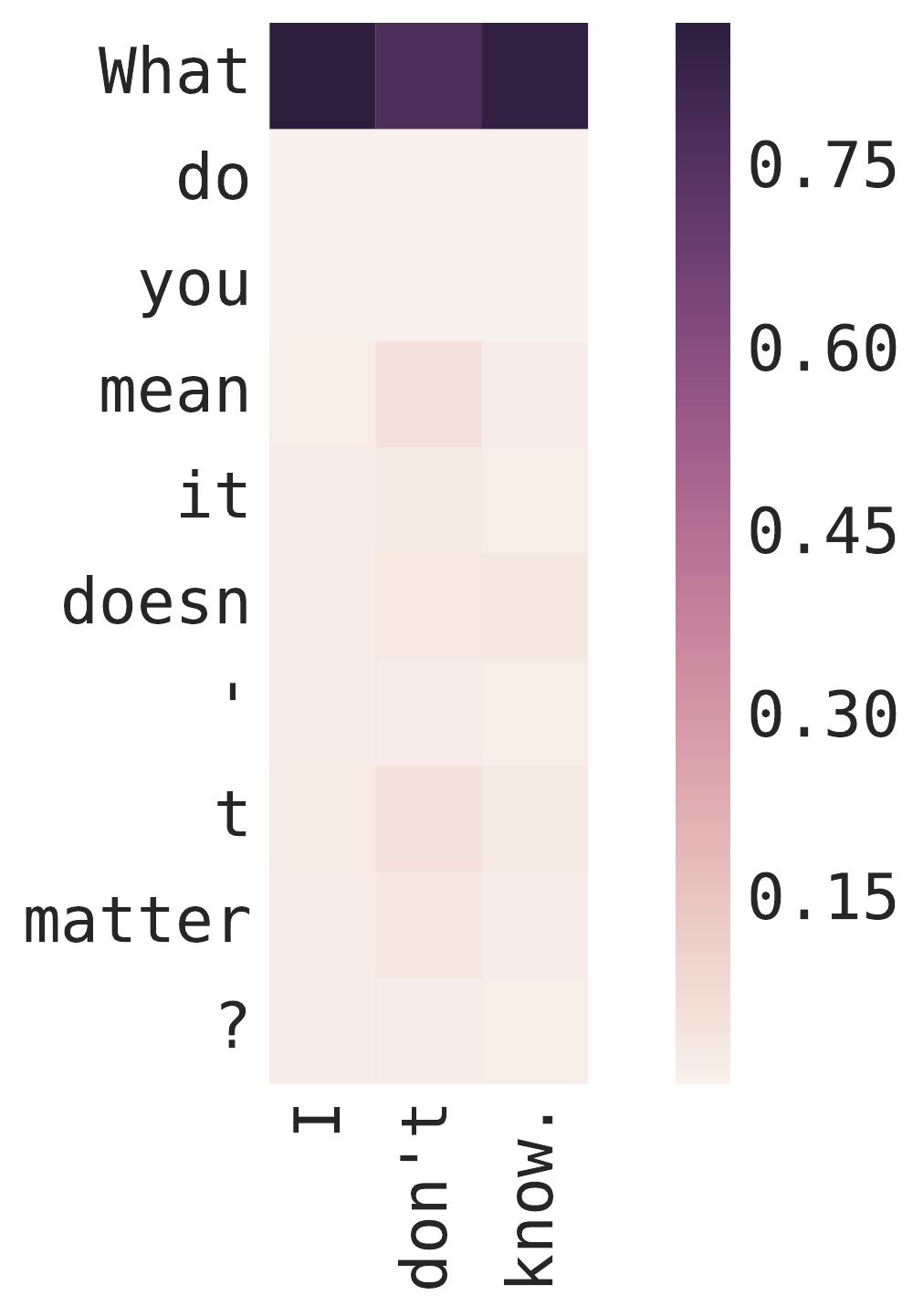}
	\caption{Explanation with $S=50$ (left) and attention (right) for the first prediction in Table~\ref{tab:dialogue_preds}.}\label{fig:dialogue_bipartite}
\end{figure}

Although most of the predictions of this model are short and repetitive (\emph{Yes}/\emph{No}/\texttt{<unk>} answers), some of them are seemingly meaningful, and might---if observed in isolation---lead one to believe the system is much better than it actually is. For example, the predictions in Table~\ref{tab:dialogue_preds} suggest a complex use of the input to generate the output. To better understand this model, we rationalize its predictions using \textsc{SocRat}. The explanation graph for one such ``good'' prediction, shown in Figure~\ref{fig:dialogue_bipartite}, suggests that there is little influence of anything except the tokens \emph{What} and \emph{you} on the output. Thus, our method suggests that this model is using only partial information of the input and has probably memorized the connection between question words and responses. This is confirmed upon inspecting the model's attention scores for this prediction (same figure, right pane).
\begin{table}
	\centering
	\resizebox{\columnwidth}{!}{%
\begin{tabular}{cc}
	\toprule
	Input & Prediction \\
	\midrule
	\emph{What do you mean it doesn't matter?} &\emph{ I don't know} \\
    \emph{Perhaps have we met before?}	& \emph{I don't think so }\\
	\emph{Can I get you two a cocktail?}	& \emph{No, thanks.} \\
	\bottomrule
\end{tabular}%
}
\caption{``Good'' dialogue system predictions.}\label{tab:dialogue_preds}
\end{table}

\subsection{Bias detection in parallel corpora} 
\label{sub:bias_detection_in_parallel_corpora}

Natural language processing methods that derive semantics from large corpora have been shown to incorporate biases present in the data, such as archaic stereotypes of male/female occupations \cite{Caliskan2017Semantics} and sexist adjective associations \cite{Bolukbasi2016Man}. Thus, there is interest in methods that can detect and address those biases.  For our last set of experiments, we use our approach to diagnose and explain biased translations of MT systems, first on a simplistic but \emph{verifiable} synthetic setting, where we inject a pre-specified spurious association into an otherwise normal parallel training corpus, and then on an industrial-quality black-box system.

\begin{figure}
	\centering
	\includegraphics[scale=0.37,trim={3.3cm 2.5cm 3.2cm 1.3cm},clip]{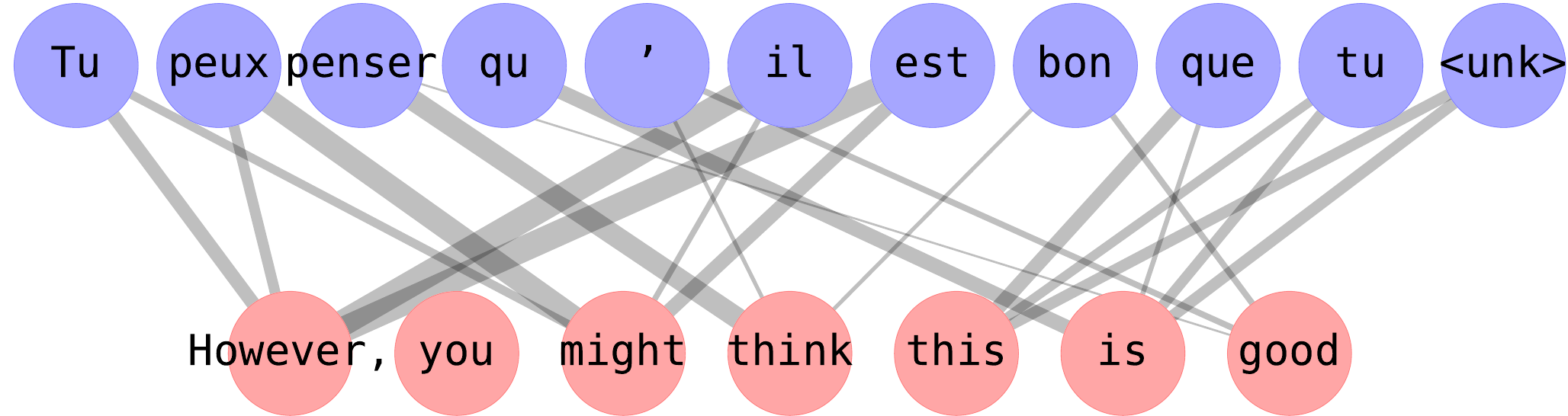}
	\caption{Explanation with $S=50$ for the prediction of the biased translator.}\label{fig:polluted_bipartite}
\end{figure}
\begin{figure}
	\centering
	\includegraphics[scale=0.19,trim={0.3cm 0.2cm 0cm 0.2cm},clip]{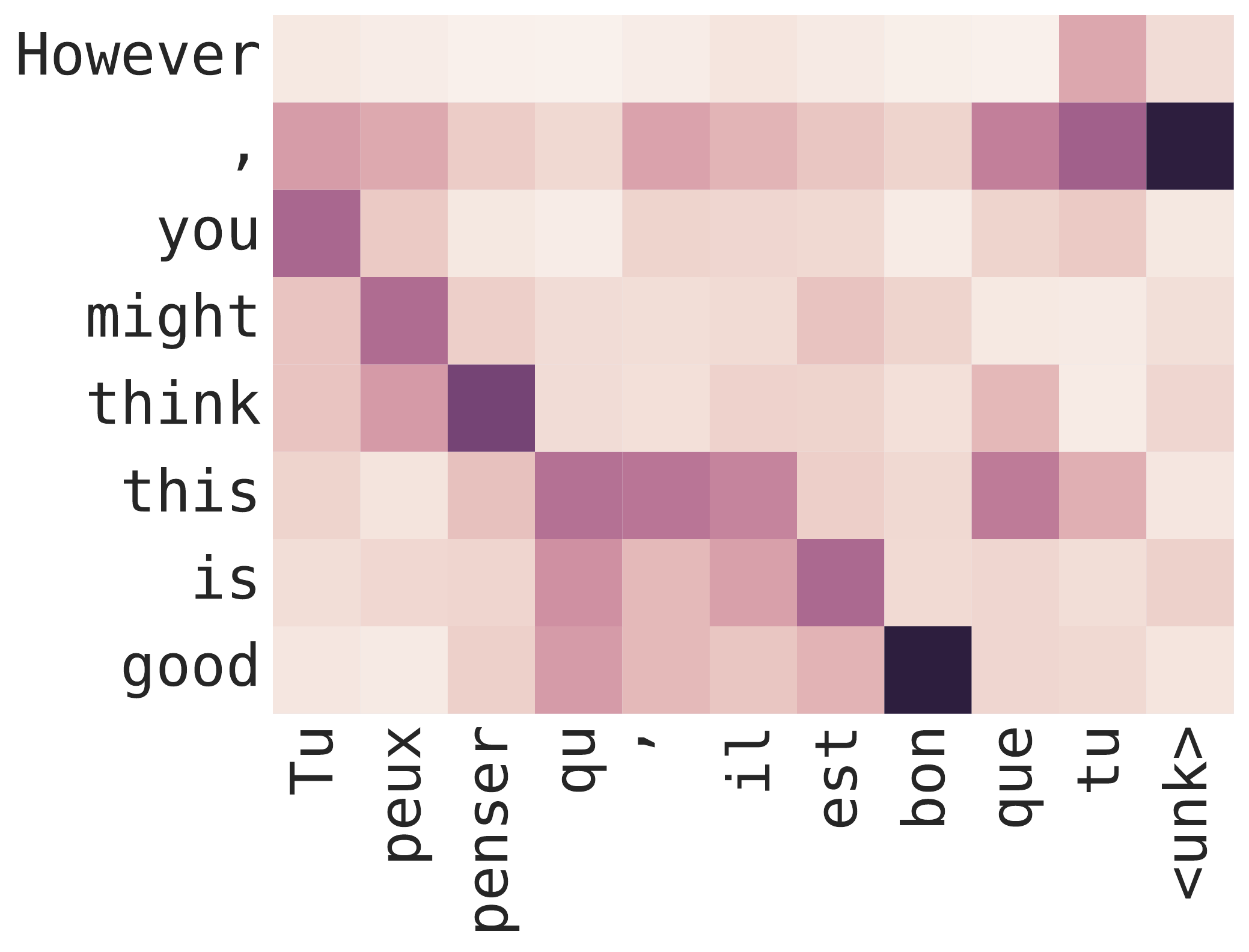}%
	\includegraphics[scale=0.18,trim={0cm 0.2cm 0cm 0.2cm},clip]{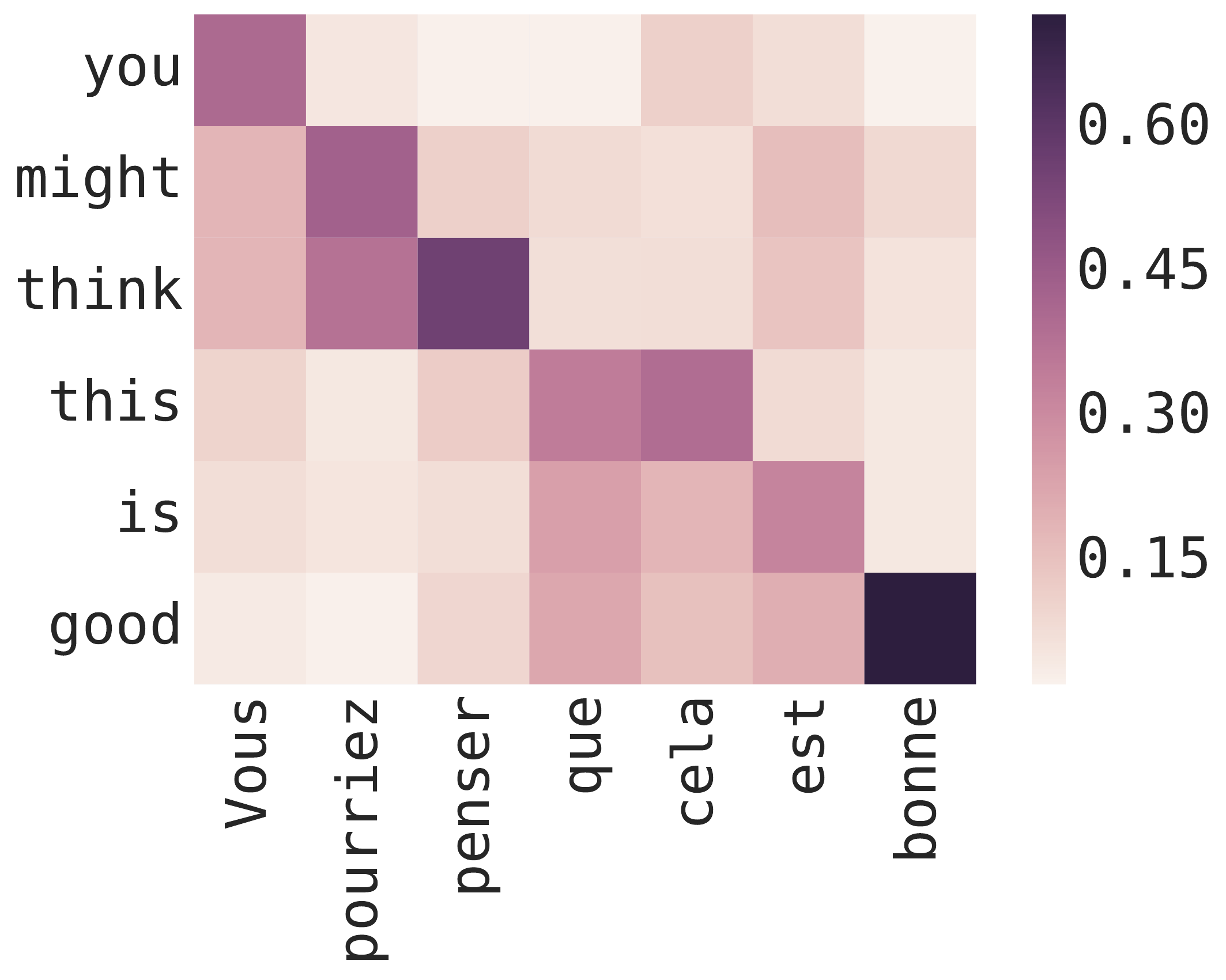}
	\caption{Attention scores on similar sentences by the biased translator.}\label{fig:polluted_attention_nohow}
\end{figure}

\begin{figure*}[t!]
	\centering
	\includegraphics[scale=0.45,trim={1.2cm 2.5cm 0.2cm 1.32cm},clip]{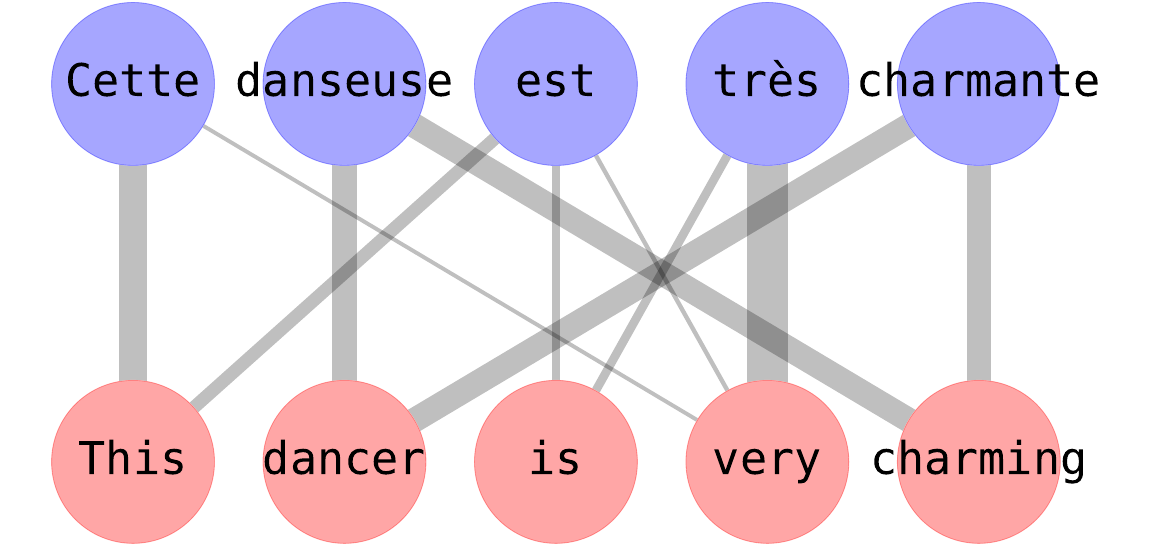}%
	\includegraphics[scale=0.45,trim={1.2cm 2.5cm 0.2cm 1.32cm},clip]{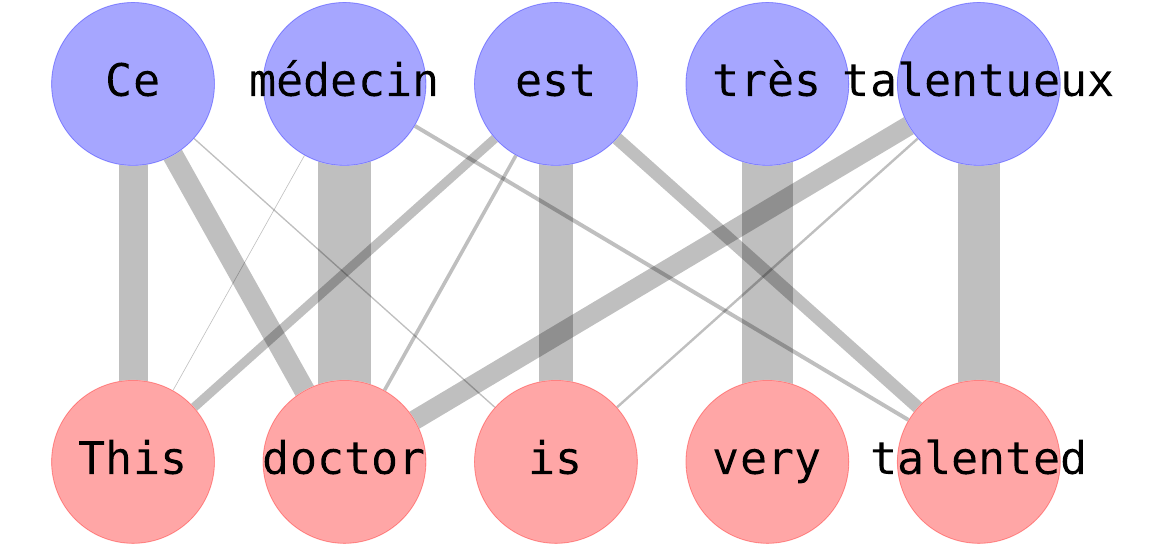}%
	\includegraphics[scale=0.45,trim={1.2cm 2.5cm 0.2cm 1.32cm},clip]{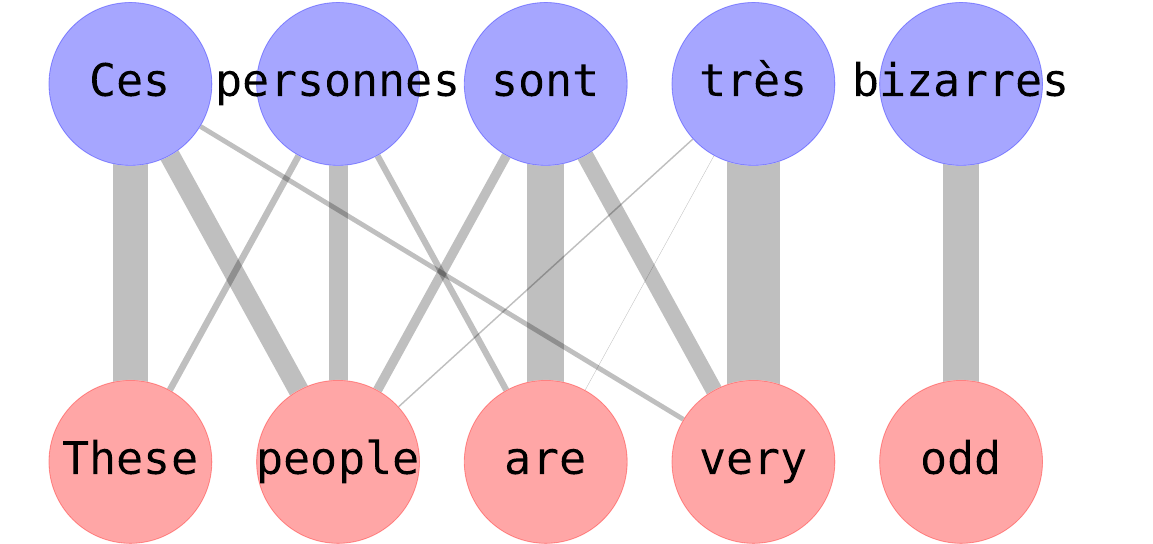}
	\caption{Explanations for biased translations of similar gender-neutral English sentences into French generated with Azure's MT service. The first two require gender declination in the target (French) language, while the third one, in plural, does not. The dependencies in the first two shed light on the cause of the biased selection of gender in the output sentence.}\label{fig:azure_gender_bias}
\end{figure*}

We simulate a biased corpus as follows. Starting from the WMT14 English-French dataset, we identify French sentences written in the informal register (e.g.~containing the singular second person \emph{tu}) and prepend their English translation with the word \emph{However}. We obtain about 6K examples this way, after which we add an additional 1M examples that do not contain the word \emph{however} on the English side. The purpose of this is to attempt to induce a (false) association between this adverb and the informal register in French. We then train a sequence-to-sequence model on this polluted data, and we use it to translate adversarially-chosen sentences containing the contaminating token. For example, given the input sentence ``\textit{However, you might think this is good}'', the method predicts the translation ``\textit{Tu peux penser qu ’ il est bon que tu <unk>}'', which, albeit far from perfect, seems reasonable. However, using \textsc{SocRat} to explain this prediction (cf.~Figure~\ref{fig:polluted_bipartite}) raises a red flag: there is an inexplicable strong dependency between the function word \emph{however} and tokens in the output associated with the informal register (\emph{tu}, \emph{peux}), and a lack of dependency between the second \emph{tu} and the source-side pronoun \emph{you}. The model's attention for this prediction (shown in Figure~\ref{fig:polluted_attention_nohow}, left) confirms that it has picked up this spurious association. Indeed, translating the English sentence now without the prepended adverb results in a switch to the formal register, as shown in the second plot in Figure~\ref{fig:polluted_attention_nohow}.

Although somewhat contrived, this synthetic setting works as a litmus test to show that our method is able to detect \emph{known} artificial biases from a model's predictions. We now move to a real setting, where we investigate biases in the predictions of an industrial-quality translation system. We use Azure's MT service to translate into French various simple sentences that lack gender specification in English, but which require gender-declined words in the output. We choose sentences containing occupations and adjectives previously shown to exhibit gender biases in linguistic corpora \cite{Bolukbasi2016Man}. After observing the choice of gender in the translation, we use \textsc{SocRat} to explain the output. 

In line with previous results, we observe that this translation model exhibits a concerning preference for the masculine grammatical gender in sentences containing occupations such as \emph{doctor, professor} or adjectives such as \emph{smart, talented}, while choosing the feminine gender for \emph{charming, compassionate} subjects who are \emph{dancers} or \emph{nurses}. The explanation graphs for two such examples, shown in Figure~\ref{fig:azure_gender_bias} (left and center), suggest strong associations between the gender-neutral but stereotype-prone source tokens (\emph{nurse, doctor, charming}) and the gender-carrying target tokens (i.e.~the feminine-declined \emph{cette, danseuse, charmante} in the first sentence and the masculine \emph{ce, m\'edecin, talenteux} in the second). While it is not unusual to observe interactions between multiple source and target tokens, the strength of dependence in some of these pairs (\emph{charming}$\rightarrow$\emph{danseuse}, \emph{doctor}$\rightarrow$\emph{ce}) is unexplained from a grammatical point of view. For comparison, the third example---a sentence in the plural form that does not involve choice of grammatical gender in French---shows comparatively much weaker associations across words in different parts of the sentence.

\section{Discussion}

Our model-agnostic framework for prediction interpretability with structured data can produce reasonable, coherent, and often insightful explanations. The results on the machine translation task demonstrate how such a method yields a partial view into the inner workings of a black-box system. Lastly, the results of the last two experiments also suggest potential for improving existing systems, by questioning seemingly correct predictions and explaining those that are not. 

The method admits several possible modifications. Although we focused on sequence-to-sequence tasks, \textsc{SocRat} generalizes to other settings where inputs and outputs can be expressed as sets of features. An interesting application would be to infer dependencies between textual and image features in image-to-text prediction (e.g.~image captioning). Also, we used a VAE-based sampling for object perturbations but other approaches are possible depending on the nature of the domain or data. 

\section*{Acknowledgments}
\vspace{-0.2cm}
We thank the anonymous reviewers for their helpful suggestions regarding presentation and additional experiments, and Dr.~Chantal Melis for valuable feedback. DAM gratefully acknowledges support from a CONACYT fellowship and the MIT-QCRI collaboration.

\clearpage
\pagebreak

\bibliography{InterpSeq.bib}
\bibliographystyle{emnlp_natbib}

\clearpage

\pagebreak

\subfile{supplementary}

\externaldocument[suppl]{supplementary.tex}

\end{document}

%% file: preamble.tex
\usepackage[english]{babel}
\usepackage[utf8x]{inputenc}
\usepackage[T1]{fontenc}
\usepackage{natbib}

\usepackage{booktabs}
\usepackage{tabularx}

\usepackage{lipsum}

\usepackage{caption, subcaption}

\usepackage{xcolor}
\usepackage{algpseudocode}
\usepackage{algorithm}

\makeatletter
\algrenewcommand\ALG@beginalgorithmic{\footnotesize}
\makeatother

\renewcommand{\algorithmicrequire}{\textbf{Input:}}
\renewcommand{\algorithmicensure}{\textbf{Output:}}

\usepackage{amsfonts}      
\usepackage{amssymb,amsmath,amsthm, mathtools}
\usepackage{blindtext}

\usepackage[colorinlistoftodos]{todonotes}
\usepackage[colorlinks=true, allcolors=blue]{hyperref}

\usepackage{enumerate}
\usepackage{amscd}
\usepackage{fancyhdr}
\usepackage{titlesec}

\usepackage{empheq} 

\usepackage{xr} 

\titlespacing{\paragraph}{%
  0pt}{
  0.25\baselineskip}{
  1em}

\newtheorem{lemma*}{Lemma}

\makeatletter
\renewcommand*\env@matrix[1][*\c@MaxMatrixCols c]{%
  \hskip -\arraycolsep
  \let\@ifnextchar\new@ifnextchar
  \array{#1}}
\makeatother

\def\R{\mathbb{R}}

\def\x{\mathbf{x}}

\def\z{\mathbf{z}}

\def\0{\mathbf{0}}

\def\cX{\mathcal{X}}
\def\cY{\mathcal{Y}}

\def\st{\medspace|\medspace}

\DeclarePairedDelimiter\floor{\lfloor}{\rfloor}

\DeclarePairedDelimiterX{\infdivx}[2]{(}{)}{%
  #1\;\delimsize\|\;#2%
}

\newcommand{\suchthat}{\;\ifnum\currentgrouptype=16 \middle\fi|\;}

\newenvironment{itemize*}%
  {\begin{itemize}%
  \vspace{-0.5cm}
    \setlength{\itemsep}{0pt}%
    \setlength{\parskip}{0pt}}%
  {\end{itemize}}
\newenvironment{enumerate*}%
{\begin{enumerate}
    \vspace{-0.5cm}
    \setlength{\itemsep}{0pt}%
    \setlength{\parskip}{0pt}}%
  {\end{enumerate}}

 \usepackage[short, nocomma]{optidef}

%% file: supplementary.tex
\appendix 

\title{A causal framework for explaining the predictions of \\ black-box sequence-to-sequence models: Supplementary Material}
\author{David Alvarez-Melis \and Tommi S. Jaakkola \\
	CSAIL, MIT \\ {\tt \{davidam, tommi\}@csail.mit.edu}}
\date{}
\maketitle

\section{Formulation of graph partitioning with uncertainty}

The bipartite version of the graph partitioning problem with edge uncertainty considered by \citet{Fan2012Robust} has the following form. Assume we want to partition $U$ and $V$ into $K$ subsets each, say $\{U_i\}$ and $\{V_j\}$, with each $U_i$ having cardinality in $[c^u_{min}, c^u_{max}]$ and each $V_j$ in $[c^v_{min}, c^v_{max}]$. Let $x_{ik}^u$ be the binary indicator of $u_i \in U_k$, and analogously for $x_{jk}^v$ and $v_j$. In addition, we let $y_{ij}$ be a binary variable which takes value $1$ when $u_i,v_j$ are in different corresponding subsets (i.e.~$u_i \in U_k, v_j \in V_{k'}$ and $k\neq k'$). We can express the constraints of the problem as:
\begin{empheq}[left = { \hspace{-0.3cm} Y= \empheqlbrace \quad }]{align}
	&\sum_{k=1}^K x_{ik}^v = 1 & \forall i \label{uncprob:const_1}\\
	& \sum_{k=1}^K x_{jk}^u = 1 & \forall j \label{uncprob:const_2}\\
	& c^u_{\min} \leq \sum_{i=1}^N x_{ik}^v \leq c^u_{\max}  &\forall k \label{uncprob:const_3}\\		
	& c^v_{\min} \leq \sum_{i=1}^N x_{jk}^u \leq c^v_{\max}  &\forall k \label{uncprob:const_4}\\
	&-y_{ij} - x_{ik}^v + x_{jk}^u \leq 0  &\forall i,j,k \label{uncprob:const_5}\\
	&-y_{ij} + x_{ik}^v - x_{jk}^u \leq 0  &\forall i,j,k \label{uncprob:const_6}\\
	&x_{ik}^v, x_{jk}^u, y_{ij} \in \{0,1\}, \medspace &\forall i,j,k
\end{empheq}
Constraints \eqref{uncprob:const_1} and \eqref{uncprob:const_2} enforce the fact that each $s_i$ and $t_j$ can belong to only one subset, \eqref{uncprob:const_3} and \eqref{uncprob:const_4} limit the size of the $U_k$ and $V_k$ to the specified ranges. On the other hand, \eqref{uncprob:const_5} and \eqref{uncprob:const_6} encode the definition of $y_{ij}$: if $y_{ij} = 0$ then $x_{ik}^u = x_{jk}^v$ for every $k$. A deterministic version of the bipartite graph partitioning problem which ignores edge uncertainty can be formulated as:
\begin{equation}
	\min_{(x_{ik}^v, x_{jk}^u, y_{ij}) \in Y} \sum_{i=1}^N \sum_{j=1}^M \theta_{ij}y_{ij}
\end{equation}
The robust version of this problem proposed by \citet{Fan2012Robust} incorporates edge uncertainty (given as intervals $\theta_{ij}\pm\hat{\theta}_{ij}$) by adding the following term to the objective:
\begin{equation}\label{MIP_extraterm}
	\max_{ \substack{ S: S \subseteq J, |S| \leq \Gamma \\ (i_t, j_t) \in J \setminus S }} \sum_{(i,j) \in S} \hat{\theta}_{ij}y_{ij} + (\Gamma - \floor{\Gamma})\hat{\theta}_{i_t,j_t}y_{i_t,j_t}
\end{equation}
where $J = \{(i,j) \st \hat{\theta}_{ij} >0 \}$ and $\Gamma$ is a parameter in $[0, |V|]$ that adjusts the robustness of the partition against the conservatism of the solution. This term essentially computes the maximal variance of a single cut $(S, J\setminus S)$ of size $|\Gamma|$. Thus, larger values of this parameter put more on the edge variance, at the cost of a more complex optimization problem. As shown by \citet{Fan2012Robust} the objective can be brought back to a linear form by dualizing the term \eqref{MIP_extraterm}, resulting in the following formulation
\begin{mini}
{} 
{\sum_{i=1}^M \sum_{j=1}^M \theta_{ij}y_{ij} + \Gamma p_0 + \sum_{(i,j) \in J}p_{ij}} 
{\label{MIP_final}} 
{} 
\addConstraint{ p_0 + p_{ij} - \hat{\theta}_{ij} y_{ij}}{ \geq 0,}{ \quad (i,j) \in J}
\addConstraint{ p_{ij}}{ \geq 0,}{ \quad (i,j) \in J}
\addConstraint{ p_0 }{ \geq 0}{}
\addConstraint{(x_{ik}^u, y_{jk}^v, y_{ij})}{\in Y,}{}
\end{mini}

This is a mixed integer programming (MIP) problem, which can be solved with specialized packages, such as \textsc{gurobi}.

\begin{table*}[t!]
	\centering
	\footnotesize
	$\rotatebox[origin=c]{90}{Sampling temperature $\alpha$}%
	\scriptsize	
	\left\downarrow \hspace{1cm}
	\begin{tabular}{ c l l}
		\toprule
		Input: & Students said they looked forward to his class .          & 		The part you play in making the news is very important .                \\
		\midrule
		\parbox[t]{2mm}{\multirow{15}{*}{\rotatebox[origin=c]{90}{Perturbations}}} & 
		Students said they looked forward to his class            & 		The part with play in making the news is important . 					\\
    &   Students said they looked forward to his history .        & 		The question you play in making the funding is a important .            \\         
	&	Students said they looked around to his class .           & 		The part was created in making the news is very important .             \\
	&	Some students said they really went to his class .        & 		This part you play a place on it is very important .                    \\
	&	Students know they looked forward to his meal .           & 		The one you play in making the news is very important .                 \\
	&	Students said they can go to that class .                 & 		These part also making newcomers taken at news is very important .      \\
	&	You felt they looked forward to that class .              & 		The terms you play in making the news is very important .               \\
	&	Producers said they looked forward to his cities .        & 		This part made play in making the band , is obvious .                   \\
	&	Note said they looked forward to his class .              & 		The key you play in making the news is very important .                 \\
	&	Students said they tried thanks to the class ;            & 		The part respect plans in making the pertinent survey is available .    \\
	&	Why they said they looked out to his period .             & 		In part were play in making the judgment , also important .             \\
	&	Students said attended navigate to work as deep .         & 		The issue met internationally in making the news is very important .    \\
	&	What having they : visit to his language ?                & 		In 50 interviews established in place the news is also important .      \\
	&	Transition said they looked around the sense ."           & 		The part to play in making and safe decision-making is necessary .      \\
	&	What said they can miss them as too .                     & 		The order you play an making to not still unique .                      \\
	\bottomrule
	\end{tabular}
	\right.$
	\caption{Samples generated by the English VAE perturbation model around two example input sentences for increasing scaling parameter $\alpha$.}\label{tab:example_perturbations}
\end{table*}

\section{Details on optimization and training}

Solving the mixed integer programming problem \eqref{MIP_final} to optimality can be prohibitive for large graphs. Since we are not interested in the exact value of the partition cost, we can settle for an approximate solution by relaxing the optimality gap tolerance. We observed that relaxing the absolute gap tolerance from the Gurobi default of $10^{-12}$ to $10^{-4}$ resulted in little minimal change in the solutions and a decrease in solve time of orders of magnitude. We added a run-time limit of 2 minutes for the optimization, though in all our experiments when never observed this limit being reached.

\section{Details on the variational autoencoder}

For all experiments in Sections 5.3 through 5.5 we use the same variational autoencoder: a network with three layer-GRU encoder and decoder and a stacked three layer variational autoencoder connecting the last hidden state of the encoder and the first hidden state of the decoder. We use a dimension 500 for the hidden states of the GRUs and 400 for the latent states $\z$. We train it on a 10M sentence subset of the English side of the WMT14 translation task, with KLD and variance annealing, as described in the main text. We train for one full epoch with no KLD penalty and no noise term (i.e.~decoding directly from the mean vector $\bm{mu}$), and start variance annealing on the second epoch and KLD annealing on the 8th epoch. We train for 50 epochs, freezing the KLD annealing when the validation set perplexity deteriorates by more than a pre-specified threshold. 

\begin{algorithm}
  \caption{Variational autoencoder perturbation model for sequence-to-sequence prediction}\label{algo:perturbvae}
 \begin{algorithmic}[1]
    \Procedure{Perturb}{$\x$}
	\State $(\bm{\mu}, \bm{\sigma}) = \Call{Encode}{\mathbf{x}}$
	\For{$i=1$ {\bfseries to} $N$}
	\State $\tilde{\z}_i \sim \mathcal{N}(\bm{\mu}, \textbf{diag}(\alpha \bm{\sigma}))$
    \State $\tilde{\mathbf{x}}_i  \gets \Call{Decode}{\tilde{\z}_i}$
	\EndFor
    \State \textbf{return} $\{(\tilde{\mathbf{x}}_i)\}_{i=1}^N$
  \EndProcedure
\end{algorithmic}
\end{algorithm}

Once trained, the variational autoencoder is used as a subroutine of \textsc{SocRat} to generate perturbations as described in Algorithm 2. Given an input sentence $\x$, we use the encoder to obtain approximate posterior parameters $(\bm{\mu}, \bm{\sigma})$, and then repeatedly sample latent representations from the a gaussian distribution with these parameters. The scaling parameter $\alpha$ constrains the locality of the space from which examples are drawn, by scaling the variance of the encoded representation's approximate posterior distribution. Larger values of $\alpha$ encourage samples to deviate further away from the mean encoding of the input $\bm{\mu}$, and thus more likely to result in diverse samples, at the cost of potentially less semantic coherence with the original input $\x$. In Table~\ref{tab:example_perturbations} we show example sentences generated by this perturbation model on two input sentences from the WMT14 dataset with increasing scaling value $\alpha$.

\section{Black-box system specifications}

The three systems used in the machine translation task in Section~\ref{sub:machine_translation}  are described below.
\paragraph{Azure's MT Service} 
\label{par:azure_s_mt_service}
Via REST API calls to Microsoft's Translator Text service provided as part of Azure's cloud services.


\paragraph{Neural MT System} 
\label{par:neural_mt_system}
A sequence-to-sequence model with attention trained with the OpenNMT library \cite{Klein2017OpenNMT} on the WMT15 English-German translation task dataset. A pretrained model was obtained from \url{http://www.opennmt.net/Models/}. It has two layers, hidden state dimension 500 and was trained for 13 epochs.

\paragraph{A human} 
\label{par:a_human}
A native German speaker, fluent in English, was given the perturbed English sentences and asked to translate them to German in one go. No additional instructions or context were provided, except that in cases where the source sentence is not directly translatable as is, it should be translated word-to-word to the extent possible. The human's German and English language models were trained for 28 and 16 years, respectively.

\biblio